\documentclass{article} %
\usepackage{colm2024_conference}

\usepackage{microtype}
\usepackage{url}
\usepackage{tabularx} %

\usepackage{wrapfig}
\usepackage{array}    %
\usepackage{booktabs}
\usepackage{multicol} %
\usepackage{amsthm} %
\usepackage{appendix}
\usepackage{enumitem}
\usepackage[linesnumbered,ruled,vlined]{algorithm2e}
\usepackage{subfigure}
\usepackage{multirow}
\usepackage{balance}
\usepackage[normalem]{ulem}

\usepackage{amsfonts,amssymb}
\usepackage{amsmath,bm}
\usepackage{threeparttable}

\usepackage{color}
\usepackage{soul} 
\usepackage{float}
\usepackage{makecell} 
\usepackage{longtable} %
\usepackage{ltablex} %

\usepackage{pifont}
\usepackage{graphicx}
\usepackage{fancyhdr}
\usepackage{longtable}
\usepackage{subcaption}

\theoremstyle{plain}
\newtheorem{theorem}{Theorem}[section]

\definecolor{hiddendraw}{RGB}{205, 44, 36}
\definecolor{hidden-blue}{RGB}{194,232,247}
\definecolor{hidden-orange}{RGB}{243,202,120}
\definecolor{hidden-yellow}{RGB}{242,244,193}
\definecolor{uclablue}{rgb}{0.15, 0.45, 0.68}
\usepackage{hyperref}
\hypersetup{
    breaklinks,
    colorlinks=true,
    citecolor=uclablue,
    urlcolor=uclablue,
}

\title{Evaluating the Factuality of Large Language Models using Large-Scale Knowledge Graphs}

\colmfinalcopy
\author{Xiaoze Liu\textsuperscript{$\clubsuit$}, Feijie Wu\textsuperscript{$\clubsuit$}, Tianyang Xu\textsuperscript{$\clubsuit$}, Zhuo Chen\textsuperscript{$\heartsuit$},Yichi Zhang\textsuperscript{$\heartsuit$}, \\ \textbf{Xiaoqian Wang\textsuperscript{$\clubsuit$}, Jing Gao\textsuperscript{$\clubsuit$}}\\
\textsuperscript{$\clubsuit$} Purdue University,
West Lafayette, IN 47907, USA \\
\textsuperscript{$\heartsuit$} Zhejiang University, Hangzhou, China\\
\texttt{\{xiaoze, wu1977, xu1868, joywang, jinggao\}@purdue.edu} \\
\texttt{\{zhuo.chen, zhangyichi2022\}@zju.edu.cn}
}

\newcommand*{\GraphEval}[0]{\textsf{GraphEval}}
\newcommand*{\Truthful}[0]{\textit{Truthful}}
\newcommand*{\Informative}[0]{\textit{Informative}}
\newcommand*{\Correct}[0]{\textit{Correct}}
\newcommand{\removevspace}[1]{}

\theoremstyle{definition}
\newtheorem{example}{Example}[section] 
\renewcommand{\paragraph}[1]{\vspace{-1mm}
    \noindent\textbf{#1.}}

\newcommand{\rmnum}[1]{\romannumeral #1}
\newcommand{\Rmnum}[1]{\expandafter\@slowromancap\romannumeral #1@}

\newcommand{\cz}[1]{{#1}}

\newcommand{\feijie}[1]{#1}

\begin{document}

\maketitle

\begin{abstract}
The advent of Large Language Models (LLMs) has significantly transformed the AI landscape, enhancing machine learning and AI capabilities. Factuality issue is a critical concern for LLMs, as they may generate factually incorrect responses.
In this paper, we propose \GraphEval{} to evaluate an LLM's performance using a substantially large test dataset. Specifically, the test dataset is retrieved from a large knowledge graph with more than 10 million facts without expensive human efforts. Unlike conventional methods that evaluate LLMs based on generated responses, \GraphEval{} streamlines the evaluation process by creating a judge model to estimate the correctness of the answers given by the LLM. Our experiments demonstrate that the judge model's factuality assessment aligns closely with the correctness of the LLM's generated outputs, while also substantially reducing evaluation costs.  Besides, our findings offer valuable insights into LLM performance across different metrics and highlight the potential for future improvements in ensuring the factual integrity of LLM outputs. The code is publicly available at \url{https://github.com/xz-liu/GraphEval}.

\end{abstract}

\section{Introduction}

The rapid progress of Large Language Models (LLMs) has markedly boosted artificial intelligence and machine learning due to their strong contextual text generation capabilities.
\feijie{Despite these groundbreaking advancements, recent works~\citep{EvaluationSurvey, wang2023evaluating} have highlighted the significance of LLMs evaluation.}
\cz{LLMs are prone to producing seemingly authentic yet factually inaccurate responses, a phenomenon known as hallucination~\citep{wang2023survey}. Such errors may stem from outdated or incorrect data during training or the model's learned associations, impacting its reliability~\citep{DBLP:journals/corr/abs-2402-05391}.}
\feijie{The evaluation, therefore, helps identify instances of hallucination and understand the LLM's ability to generate coherent and contextually relevant text, i.e., factuality of LLM outputs.}

\feijie{Recent efforts have been put into the factuality evaluation of LLM. For example, \citet{chen2023felm} introduce FELM, a benchmark comprising diverse factual samples across various domains. Moreover, \citet{tian2023finetuning} and \citet{feng-etal-2023-factkb} utilize external tools (e.g., search engines and a well-trained factual LLM) to estimate the factuality of the generated texts. Representing structured knowledge related to real-world objects, Knowledge Graphs (KGs) \citep{auer2007dbpedia, bollacker2008freebase, suchanek2007yago, carlson2010toward} have gained prominence for LLM factuality assessments. They primarily originate from Wikipedia, encapsulate factual information for AI tasks, and form the knowledge base with datasets like Natural Questions \citep{NaturalQuestions}. Several studies \citep{sun2023head,liang2023holistic} have focused on creating benchmarks from knowledge sources by posing factual questions derived from triples in} \cz{KGs}. %
These methods, as depicted in the left part of Figure \ref{fig:intro_demo}, either (i) sample subgraphs from large KGs or (ii) extract a subset of knowledge from text documents to construct multiple-choice or text question-answer pairs. Those question pairs are then posed to LLMs to assess their factuality.

However, the above-mentioned methods or evaluation strategies face challenges in comprehensively evaluating the factuality of LLMs. 
{\it Firstly}, \cz{the scope of evaluation data is often limited or incomplete, focusing predominantly on specific domains.}
\cz{This limitation restricts the evaluation's breadth and undermines its applicability across various contexts, failing to cover the wide range of topics LLMs are expected to handle.}
The specialized nature of these datasets means that the evaluation may not accurately reflect the model's performance in generating factual content across a broader spectrum of subjects. {\it Secondly}, the process of factuality evaluation itself is inherently time-consuming and costly. It necessitates that an LLM generate full texts, which must then be meticulously assessed for accuracy and reliability. 
\cz{This comprehensive generation and detailed review demand significant computational resources and extensive human effort~\citep{wang2023evaluating} for validation.}
As a result, the process becomes less feasible for regular or large-scale applications, limiting the frequency and scope of 
\cz{practical evaluations}.
{\it Lastly}, due to the limited size of the evaluation data, there may be biases in the benchmarks~\citep{gallegos2023bias}, or risks of test data leakage \citep{zhou2023dont}, which might compromise the validity of the evaluations.
Together, these challenges underscore the need for more scalable, efficient, and domain-agnostic approaches to evaluating the factuality of LLM-generated texts.

To this end, we propose \GraphEval{}, which consists of two novel features in terms of the design, as presented on the right side of Figure \ref{fig:intro_demo}. First, we utilize \cz{KGs} that encapsulate factual information sourced from verifiable content like Wikipedia. With the KG, millions of prompts can be automatically generated, leading to a significant saving of human efforts in labeling the ground truth. From the data perspective, the KG gives a more diversified and comprehensive evaluation of the LLMs' factuality. Second, the proposed method efficiently reduces the computation costs and speeds up the evaluation process. Specifically, we incorporate a highly reliable and lightweight judge model to decide whether an LLM can generate an accurate response to a designated question. 
Instead of generating the full text, the judge model returns three options (i.e., True, False, and I don't know) to simulate LLMs' responses to a given prompt. To ensure the reliability of the simulated results, the judge model is trained based on a few question-answer pairs, where the questions are sampled from KGs and the answers are generated by the target LLM. As a result, the judge model can serve as a replacement for the factuality evaluation of the facts extracted from large-scale KGs.   In summary, our contributions are as follows:

\begin{figure}[t]
    \centering 
    \includegraphics[width=.97\textwidth] {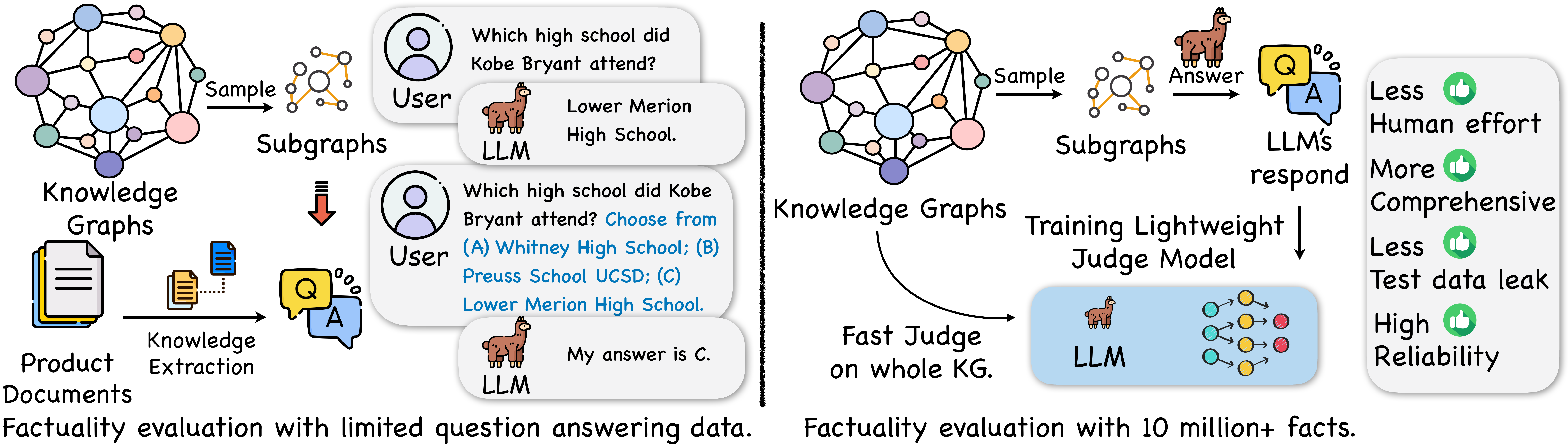} 
    \vspace{-2mm} 
    \caption{\feijie{Existing works compared to the proposed \GraphEval{} on factuality evaluation.}} 
    \vspace{-4mm}
    \label{fig:intro_demo}
\end{figure}  

\begin{itemize}[topsep=0pt,itemsep=0pt,parsep=0pt,partopsep=0pt,leftmargin=*]
    \item We propose \GraphEval{}, a large-scale evaluation framework that assesses the factuality of LLMs using KGs. \GraphEval{} evaluates the factuality of LLMs using the entire KGs, providing a more diversified and comprehensive evaluation of the LLMs' factuality.
    \item We introduce a judge model to assist with the evaluation process, which reduces the computational cost and enhances the efficiency of the evaluation. We also give a theoretical analysis of the judge model to demonstrate its validity.
    \item We conduct extensive experiments on a large-scale KG, i.e., DBpedia, to demonstrate the effectiveness and efficiency of \GraphEval{} in evaluating the factuality of LLMs.
    \item We provide an in-depth analysis of the LLM's performance on the KGs, including the LLM's performance with respect to relation types, head entity types, tail entity types, and the relation of LLM performance to degree and pageviews. 
\end{itemize}

\vspace{-2mm}
\section{Related Work}
\vspace{-2mm}
\paragraph{\cz{Factuality Issue of LLMs}}
Factuality issue~\citep{wang2023survey, zhang2023siren}, is the issue that LLMs may produce content inconsistent with established facts. As outlined in~\cite{wang2023survey}, this issue may be due to: \cz{\textbf{\textit{(\rmnum{1})}}} LLMs lacking expertise in specific domains~\citep{ScienceQA,liu2024chipnemo,9499743,bolton2024biomedlm}; \cz{\textbf{\textit{(\rmnum{2})}}} LLMs' unawareness of recent developments or changes~\citep{TempQuestions,yao2023editing}; \cz{\textbf{\textit{(\rmnum{3})}}} LLMs not retaining~\citep{wang2023evaluating,TQ,NaturalQuestions} or forgetting~\citep{goodfellow2015empirical,kotha2023understanding,wang2022preserving, chen2020recall, zhai2023investigating} knowledge from its training corpus; and \cz{\textbf{\textit{(\rmnum{4})}}} LLMs failing to reason with the knowledge they possess~\citep{liu2023we, berglund2023reversal,tan2023chatgpt}. The factuality issue has been addressed by various works, by incorporating Retrieval Augmented Generation (RAG)~\citep{lewis2020retrieval, wang2024blendfilter}, fine-tuning~\citep{tian2023finetuning,DBLP:conf/icde/00070HCGYBZYSWY23}, \cz{reward-based alignment}~\citep{DBLP:journals/corr/abs-2311-06503}, and knowledge-enhanced models~\citep{feng-etal-2023-factkb,DBLP:journals/corr/abs-2310-06671,diao2023mixtureofdomainadapters}. 
To summarize, these approaches integrate other knowledge sources into the LLMs' training process or use them to augment the models' knowledge base, thus alleviating the factuality issue. Our work differ from them in that we evaluate the factuality of LLMs, rather than providing factuality enhancement methods.

\paragraph{\cz{Factuality Evaluation of LLMs}}
The expanding use of LLMs across various domains necessitates the assurance of their output's accuracy and reliability. 
A range of benchmarks and evaluation methodologies for assessing large language models (LLMs) are proposed.
These works primarily focus on evaluating the factuality, truthfulness, reasoning capabilities, and adaptability to new information of LLMs. MMLU~\citep{MMLU} and TruthfulQA \citep{TruthfulQA} aim to measure the factuality and truthfulness of LLMs across diverse tasks, while C-Eval \citep{C-Eval} focuses on the Chinese context, assessing models' knowledge of Chinese culture and laws. BigBench  \citep{BigBench} challenges LLMs with tasks beyond current capabilities, and HaluEval \citep{HaluEval}, SelfAware \citep{yin-etal-2023-large}, and the Pinocchio \citep{Pinocchio} benchmark explore models' propensity for generating hallucinations, their self-awareness, and their reasoning skills, respectively. REALTIMEQA  \citep{kasai2022realtimeqa} and FreshQA  \citep{vu2023freshllms} introduce dynamic benchmarks that test LLMs on current events and up-to-date knowledge. There are also works \citep{sun2023head, liang2023holistic} that propose factuality evaluation using subsets of KGs. %
However, selecting subsets of KGs to test LLMs can introduce selection bias. For example, random sampling can focus more on a few popular domains or subjects more densely connected with others, thus not showing LLMs' factuality on diversified topics.
Our work addresses this limitation by %
proposing a resource-efficient method to evaluate the factuality of LLMs which allows evaluations on whole KGs instead of subsets, thus providing a more diversified and comprehensive evaluation of the LLMs' factuality, enabling an extensive assessment of LLM's factuality and reasoning abilities in a way that existing individual benchmarks do not as they only focus on specific aspects.

\paragraph{\cz{Using KGs in LLMs}}
\cz{KGs} are structured representations of factual knowledge, typically in the form of (head, relation, tail) triples.  There are lots of efforts in constructing \citep{auer2007dbpedia,bollacker2008freebase}, integrating \citep{chen2017multilingual,DBLP:conf/semweb/0007CGPYC21, ge2021largeea,ge2021make,gao2022clusterea,liu2023unsupervised}, and reasoning~\citep{bordes2013translating,sun2019rotate,guo2024distributed,chen2024exploring} on KGs. This has made KGs an indispensable resource of factual knowledge for AI tasks. 
Currently, the most common way of integrating KGs with LLMs is using KGs as an external knowledge source to enhance LLM performance by pre-training, fine-tuning, or in-context learning~\citep{yasunaga2022deep,jiang-etal-2023-reasoninglm, zhang2024knowledge,kim2023kggpt,luo2023reasoning}. Our work is different from these works in that we use KGs to evaluate the factuality of LLMs, rather than enhancing the LLMs with KGs.

\begin{figure*}[t]
    \centering
    \includegraphics[width=\textwidth]{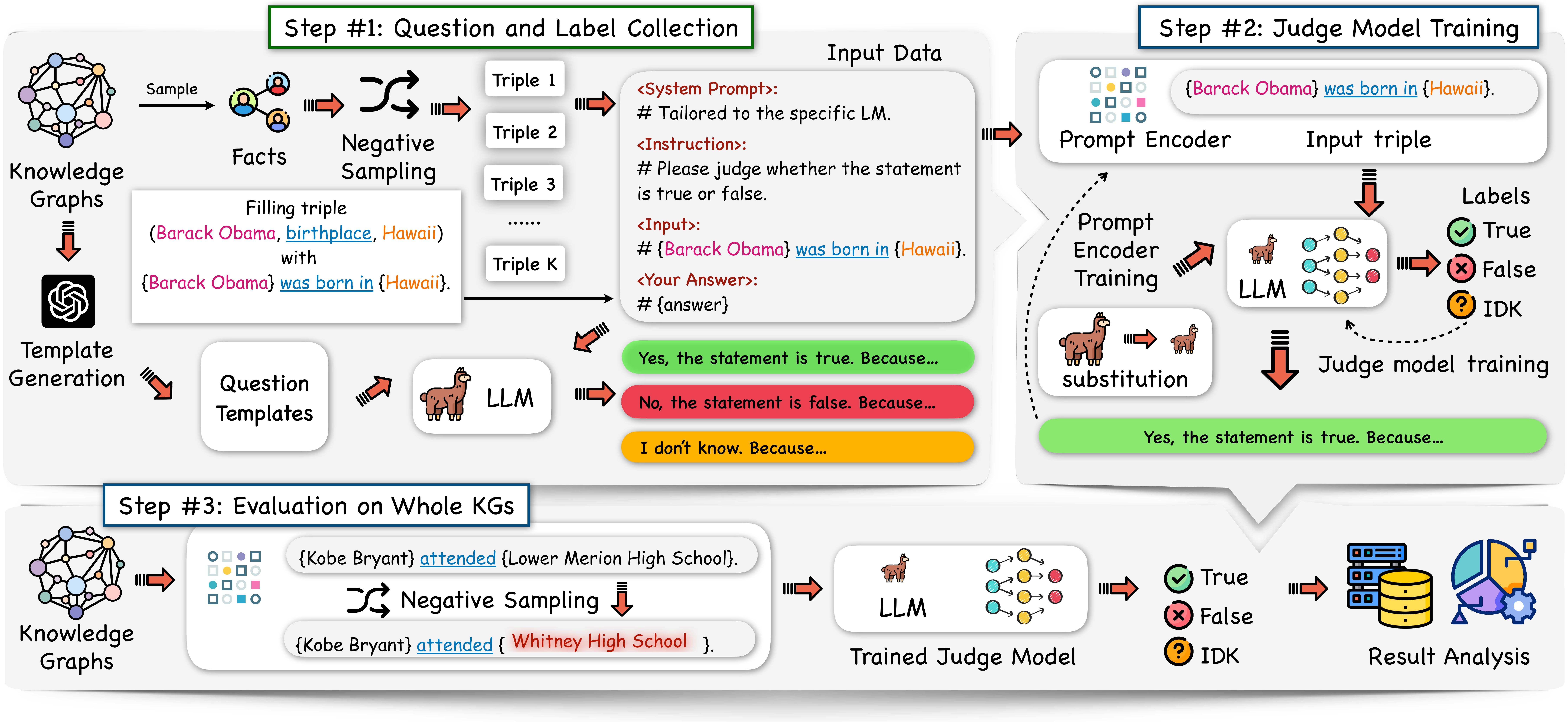} 
    \vspace{-4mm}
    \caption{Overview of the \GraphEval{} framework. \textbf{Step \#1} retrieves KG statements and collect LLM judgments on them. \textbf{Step \#2} trains the judge model which classifies LLM hidden states into three categories. \textbf{Step \#3} evaluates the LLM on all KG statements with the judge model.}
    \label{fig:framework}
\end{figure*}

\section{Method}

\vspace{-3mm}
\GraphEval{} is designed to measure the factuality of a language model in relation to a \cz{KG}. As presented in Figure \ref{fig:framework}, the proposed work is divided into three steps: 
\begin{itemize}[topsep=0pt,itemsep=0pt,parsep=0pt,partopsep=0pt,leftmargin=*]
    \item \textbf{Step 1: Question and label collection from KGs and LLMs. } \quad  The model samples triples from KGs and converts each triple into a declarative statement with GPT-4-crafted templates. To prepare versatile statements, we employ \textit{negative sampling}, where incorrect statements are intentionally generated. Afterward, those statements are posed to collect the labels answered by an LLM (i.e., Yes, No, and I don't know (IDK)).
    \item \textbf{Step 2: Judge model training.} \quad With the triples collected in the first step, we train a judge model to avoid long-generated texts and conserve computational resources. In detail, inspired by~\cite{azaria-mitchell-2023-internal}, we train a classifier with LLMs' hidden states to make a selection within the above three options. We also apply p-tuning \citep{liu2021p} to minimize the prompt/instruction size.
    \item \textbf{Step 3: Evaluation on whole KGs.} \quad Similar to the first step, we retrieve all true/false statements from KGs. Subsequently, these statements are fed into the trained judge model to estimate the factuality of LLM. This process enables a thorough and multifaceted analysis of the LLM's performance in terms of factuality, drawing from a wide range of perspectives to provide a more comprehensive and diversified evaluation.
\end{itemize}
In the following sections, we will discuss the details of each step. 

\subsection{Question and Label Collection}
\label{sec:question_generation}

\paragraph{Question Generation}
In order to evaluate the language model's ability to identify false statements, we directly construct a declarative sentence for each triple. This addresses the ineffectiveness of multiple-choice questions in our task. %
Firstly, multiple-choice prompts may cause misalignment with parametric knowledge in LLMs. Since LLMs mainly learn parametric knowledge through text data, in which knowledge facts are mostly represented as declarative sentences~\citep{weller2023according}, employing multiple-choice questions may hinder the evaluation of factuality. Secondly, multiple-choice questions have more complex labels (i.e. A, B, C, D) than declarative sentences (i.e. True, False, IDK), which can complicate the tasks for the judge model, influencing the overall effectiveness. We use an example to illustrate this.

\begin{example}
    For the triple \texttt{(Barack Obama, birthPlace, Hawaii)}, a multi-choice question can be generated as \texttt{Where was Barack Obama born?} with choices \texttt{A. Hawaii B. Chicago C. New York D. Los Angeles}. Here, for the same triple, we can also generate another multi-choice question as \texttt{Where was Barack Obama born?} with choices \texttt{A. China, B. Hawaii, C. Japan, D. Russia}. The two questions represent the same triple, but the choices are different. %
    As mentioned in the last paragraph, this can introduce complexity and potential misalignment with an LLM’s training and result in inconsistent responses. %
    
\end{example}

To address the ineffectiveness of multiple-choice questions, we propose to directly ask the LLMs whether a statement is true or not.
For instance, considering the triple \texttt{(Barack Obama, birthPlace, Hawaii)}, we can formulate a fact \texttt{Obama was born in Hawaii} by integrating the entities \texttt{Barack Obama} and \texttt{Hawaii} into the template \texttt{ \{head\} was born in \{tail\}}. %
Each template corresponds to the relation of a triple, and they are crafted to be clear and straightforward statements. GPT-4 is employed to generate these templates for all relations in the \cz{KG}. 
 These generated templates are then manually reviewed and refined to ensure their compatibility with the \cz{KG}. Then, we can ask a question to the LLMs, such as \texttt{Is the statement "Barack Obama was born in Hawaii" true or false?}.
    Here, the templates are corresponding to the relations of the triples. This is because the number of relations in the \cz{KG} is limited, while the number of triples is large. Therefore, we can use the relations to categorize the triples, and then use the templates to generate questions for each category. This can significantly reduce human labor, i.e., monitoring less than 1000 templates compared with monitoring more than 10 million triples.
See the Appendix~\ref{app:detailed_settings} for the detailed settings of the relation templates.

\paragraph{Negative sampling}
Although the declarative sentences simplify the training of the judge model, they alone are insufficient to evaluate the language model's factual accuracy. LLMs can simply answer true for every question, and still get a high accuracy. %
To address this, we introduce negative sampling, a technique commonly used in \cz{KG} completion tasks, to generate false statements. Specifically, we randomly replace one entity or relation in the original triple with another entity or relation sampled from the \cz{KG}. For example, given the triple \texttt{(Barack Obama, birthPlace, Hawaii)}, we can replace the tail entity \texttt{Hawaii} with another entity \texttt{Chicago} to form the false statement \texttt{Barack Obama was born in Chicago}. These false statements are then presented to the LLMs to evaluate their ability to identify falsehoods.
 
\vspace{-2mm}
\subsection{Judge Model}\label{method:judge}

\vspace{-2mm}
Normally, to evaluate the factual accuracy of a language model, we would generate questions from a \cz{KG} and then pose these questions to the language model. However, given the expansive nature of \cz{KG}s, it's impractical to label every generated question by the LLM. 
A more efficient approach is to use the last token logits of the LLMs as their answers. However, %
recent research has highlighted discrepancies between these logits and the model's actual text outputs~\citep{wang2024myanswer}. 
Therefore, we introduce a novel judge model to assist with this task. The judge model, initially trained on a subset of labeled questions, is then employed to label the remaining questions. Uniquely, inspired by ~\cite{azaria-mitchell-2023-internal}, the judge model utilizes the LLM's hidden state as input, as a replacement of the LLM's last layer with compressed output tokens. Specifically, three output classes are used: \textit{True}, \textit{False}, and \textit{I don't know}. The judge model is a two-layer feed-forward neural network, with a layer normalization and a ReLU activation function. 
This approach diverges from standard practices where LLMs generate answers, as here we only forward the transformer once. Consequently, this operation is significantly less resource-intensive than full answer generation, allowing the judge model to efficiently process a large number of questions with limited labeled data.

\paragraph{Efficiency}
To further enhance the judge model's efficiency, we include 2 extra components. First, we found that the instruction prefix of the LLMs is too large for the judge model to process efficiently. We thus fine-tune a {\it prompt encoder}~\citep{liu-etal-2022-p} to reduce the large input of the prompt prefix, which would be the same for all questions. 
Second, we found that our judge model, with the training process on the labeled dataset, is robust to the LLM's hidden states. %
In experiments, we observed that our judge model can seamlessly utilize hidden states from distinct LLMs without significant differences in performance. For instance, within the LLaMA 2 model family, which contains 3 models with different parameters: 7B, 13B, and 70B, we found that the judge model's performance is consistent regardless of whether the hidden states are from 7B, 13B, or 70B. %
Therefore, we can use the model with the least parameters, as a {\it substitute model} when computing the hidden states. This gives us a huge reduction in computational cost.

\paragraph{Analysis of Judge model}
In this part, we assume there are two datasets; one is for training the judge model, and the other is for evaluation, denoted by $\mathcal{D}_S$ and $\mathcal{D}_T$, respectively. As the proposed judge model leads to a triple classification task, we assume a hypothesis portfolio $h = \{h_t, h_f, h_{idk}\}$, where these three hypotheses separately predict if a sample can be correctly answered by the LLM, i.e., True, False, and IDK. In other words, the hypothesis $\hat{h} \in h$ maps an input $\mathbf{x}$ to $\{0, 1\}$, where $1$ means the input satisfies the hypothesis conditions. For a given input $\mathbf{x}$, the equality 
$h(\mathbf{x}) = h_t(\mathbf{x}) + h_f(\mathbf{x}) + h_{idk}(\mathbf{x}) = 1$
always holds because the judge model provides an only output. Define the convex loss function for a hypothesis $\hat{h} \in h$ to be $$L_{\mathcal{D}}(\hat{h}) = \sum_{(\mathbf{x}, y) \in \mathcal{D}} |\hat{h}(\mathbf{x}) - \boldsymbol{1}_{\hat{h}}(y)|$$, where $\boldsymbol{1}_{\hat{h}}(y)$ indicates if the data indeed satisfies the hypothesis. Since a wrong prediction for data $(\mathbf{x}, y)$ results in $\sum_{\hat{h} \in h} |\hat{h}(\mathbf{x}) - \boldsymbol{1}_{\hat{h}}(y)| = 2$, we define the misclassification rate as 
$$L_{\mathcal{D}} \left(h\right) = \frac{1}{2} \left(L_{\mathcal{D}} \left(h_t\right) + L_{\mathcal{D}} \left(h_f\right) + L_{\mathcal{D}} \left(h_{idk}\right)\right) $$ 

Below is a theoretical analysis to understand the bound of the misclassification rate, which is driven by Theorem 2 of \cite{ben2010theory}.

\begin{theorem}
Let $\mathcal{H} = \{\mathcal{H}_t, \mathcal{H}_f, \mathcal{H}_{idk}\}$ be a set of hypothesis spaces of VC dimension $d$. If $\mathcal{U}_S, \mathcal{U}_T$ are the samples of size $m$ each, drawn from $\mathcal{D}_S$ and $\mathcal{D}_T$, respectively, then for any $\delta \in (0, 1)$, with probability at least $1-\delta$, for every $h \in \mathcal{H}$, we have
\begin{equation}
    L_{\mathcal{D}_T} \left(h\right) \leq L_{\mathcal{D}_S} \left(h\right) + \frac{3}{4} d_{\mathcal{H} \Delta\mathcal{H}} \left(\mathcal{U}_S, \mathcal{U}_T\right) + 6 \sqrt{\frac{2d \log \left(2m\right) + \log\left(2/\delta\right)}{m}} + \frac{1}{2} \lambda
\end{equation}
where $\lambda = \inf_{h \in \mathcal{H}} \left(L_{\mathcal{D}_S}\left(h\right) + L_{\mathcal{D}_T}\left(h\right)\right)$ is the optimal combined error, $d_{\mathcal{H} \Delta\mathcal{H}}$ measures the distribution discrepancy between two distributions. 
\end{theorem}
\feijie{The above theorem provides insights for the generalization bound of the judge model. The bound is associated with the discrepancy between training data $\mathcal{D}_S$ and evaluation data $\mathcal{D}_T$, and the discrepancy can be measured by drawing samples from both training and evaluation datasets for an equivalent size. Moreover, the bound is affected by the optimal hypothesis over all the data, i.e., $\mathcal{D}_S \cup \mathcal{D}_T$, where a lower error leads to improved performance of the judge model.}

\subsection{Evaluation}

For evaluating the LLM's performance, we consider \textit{Correctness}, which is defined as the proportion of questions for which the LLM's response matches the true label (or false label if the question is generated from a negative triple). This captures the accuracy of the LLM in identifying correct information and distinguishing it from fabricated (negative) triples.
We also adopt the metrics of \textit{Truthfulness} and \textit{Informativeness}, as defined in \cite{TruthfulQA}. \textit{Truthfulness} refers to the likelihood of the language model (LLM) providing an honest response. A response is considered \Truthful{} if the LLM either provides the correct answer or opts for 'I don't know'. This criterion assesses the model's ability to be honest about what it knows and to admit uncertainty rather than making false statements. \textit{Informativeness} is the probability of the LLM offering any substantive information, irrespective of its accuracy. An answer is deemed \Informative{} if it is anything other than 'I don't know'. This reflects the model's capacity to provide substantial information without resorting to uncertainty or avoidance of an answer. 

When considering multiple negative triples sampled, we combine the results for all negative triples sampled from a triple $\tau$, as well as the results for their original positive triple $\tau$, to calculate the overall performance of the LLM. 
Since correctly detecting a real triple from KG is much simpler than detecting a negative triple, we want to give a max penalty to the LLM's wrong response to the real triple when designing the metric. Therefore, if a real triple is predicted as false, the LLM will score $0$ across all metrics. Then, the negative triple results are averaged to give a fine-grained evaluation of the LLM's performance.
To achieve this, for each performance metric, we define functions $\mathcal{F}$ which evaluates the LLM's response to $\tau$ and $\mathcal{F}'$  to each negative triple $\tau'$ sampled from $\tau$. The overall performance metric for $\tau$ is then calculated as:
\begin{equation}
\text{Metric}(\tau) = \max\left(0, \mathcal{F}(\tau) - \frac{1}{|\mathcal{N}(\tau)|}
\sum_{\tau' \in \mathcal{N}(\tau)} \mathcal{F}'(\tau')\right)
\end{equation}
Here, $\mathcal{N}(\tau)$ represents the set of all negative triples generated from the positive triple $\tau$. $\mathcal{F}$  and $\mathcal{F}'$ are defined as follows: \cz{\textbf{\textit{(\rmnum{1})}}} {\it Correctness.}
    $\mathcal{F}$ is defined such that it is $1$ if the judge model predicts that a real (positive) triple is True, and it is $0$ otherwise; $\mathcal{F}'$ is 0 if the judge model predicts a negative triple as False, and 1 otherwise. 
    \cz{\textbf{\textit{(\rmnum{2})}}} {\it Truthfulness.}
    When measuring \textit{Truthfulness}, $\mathcal{F}$ is set to $1$ if the judge model's prediction for the input $\tau$ is either True or IDK, and it is $0$ otherwise. Similarly, $\mathcal{F}'$ is set to $1$ if the judge model's prediction for the input $\tau'$ is True, and $0$ otherwise; and 
    \cz{\textbf{\textit{(\rmnum{3})}}} {\it Informativeness.}
    For \textit{Informativeness}, $\mathcal{F}$ is defined as $1$ if the judge model's prediction for the input $\tau$ is anything other than "I don't know", and it is $0$ otherwise. $\mathcal{F}'$ is set to $1-\mathcal{F}$ on the informativeness metric.
By applying this equation, we can systematically compute the \textit{Correctness}, \textit{Truthfulness}, and \textit{Informativeness} of an LLM's responses in a consistent and comprehensive manner, offering a detailed insight into its overall performance.

\section{Experiments}

\subsection{Experiment Setup}

\paragraph{Data}
We use DBpedia~\citep{auer2007dbpedia}, a large-scale knowledge graph constructed from Wikipedia. 
We report the statistics of the DBpedia knowledge graph in Table \ref{tab:dbpedia_stat}.  Note that there are ``dummy'' entities in DBpedia that represent a fact that is only true on a specific time period. An example is \url{https://dbpedia.org/page/Kathy\_Greenlee\_\_Tenure\_\_1}. For simplicity, we remove these dummy entities and triples related from the knowledge graph. We refer to the remaining triples as the DBpedia knowledge graph. The DBpedia knowledge graph contains 4,928,232 entities, 633 relations, and 16,915,848 triples. The average node degree of the knowledge graph is 6.80, and the density of the knowledge graph is $7.18\times 10^{-7}$. 

\begin{table}[t]
\centering\small
\begin{tabular}{c|c|c|c|c}
\toprule
\textbf{\#Entities} & \textbf{\#Relations} & \textbf{\#Triples} &  \textbf{Avg. degree} & \textbf{Density} \\ \midrule
4,928,232 & 633 & 16,915,848 & 6.80 & $7.18\times 10^{-7}$ \\ \bottomrule
\end{tabular}
\vspace{-2mm}
\caption{Statistics of the DBpedia knowledge graph.}
\label{tab:dbpedia_stat}
\vspace{-5mm}
\end{table}

\paragraph{LLMs}
In this paper, we evaluate the Meta LLaMA 2 family~\citep{touvron2023llama}, including LLaMA-2-7B, LLaMA-2-13B, and LLaMA-2-70B, and Google's Gemma~\citep{gemmateam2024gemma} including Gemma-2B and Gemma-7B. 
For each language model, we first randomly sample 2000 triples, and perform a negative sampling to obtain another 2000 negative triples. For each triple, we ask the LLM 3 times the same question, on whether the triple is true, false, or the LLM doesn't know. We use majority voting to determine the LLM's final answer. When asking, we use huggingface's pipeline with default settings and FP16 precision. 
 This is to form a labeled dataset. We randomly sample 70\% for the training set and 30\% for the validation set, then train a judge model to classify the LLM's hidden state into 3 classes: LLM correctly answering the question (True), LLM incorrectly answering the question (False), and LLM responding with I don't know (IDK). We refer to Table \ref{tab:llm_stat} for the statistics of the labeled dataset.

\paragraph{Metrics}
 For the LLM's performance, we report the estimated factuality of the LLMs on the DBpedia knowledge graph. We report the LLM's performance in terms of \textit{Truthfulness}, \textit{Informativeness}, and \textit{Correctness}. For evaluating the judge model's performance (See Appendix~\ref{app:judge_model}), we seek to maximize the similarity between the judge model's prediction and the LLM's answer. Thus, we use the common metrics Precision (P), Recall (R), and F1 score (F) to evaluate the judge model's accuracy; and the time it takes to predict to evaluate the judge model's efficiency.

\paragraph{Hyperparameter Settings}
For the judge model classifier training, we train 100 epochs with a batch size of 8. We use the Adam optimizer with a learning rate of 1e-4. We use the same settings for all the evaluated LLMs. For LLaMA 2 7B, 13B, and 70B, we use LLaMA 2 7B as the judge model's hidden state input. For Gemma 2B and 7B, we use Gemma 2B as the judge model's hidden state input.
 The judge model is trained on a server with NVIDIA A6000 GPUs.

For the training of the prompt encoder, we use the same settings for all the evaluated LLMs. To be specific, we use 20 virtual tokens, 1 transformer submodule, 12 attention heads, 12 layers, MLP as the encoder reparameterization type, 4096 as the encoder hidden size, and 2e-5 as the learning rate. We train the prompt encoder for 5 epochs with a batch size of 8. We use the Adam optimizer with a weight decay of 0.01.

 For the evaluation, we use two servers, one with NVIDIA A6000 GPUs and the other with NVIDIA A100 GPUs. For inference, we use Flash Attention 2~\citep{dao2023flashattention2} as the attention implementation, and use FP16 precision.

\subsection{LLM's Performance Analysis}

We report the estimated factuality of the LLMs on DBpedia in Table \ref{tab:llm_stat}. 
Overall, the LLaMA-2 series shows an increase in model size up to 13B, particularly in terms of balanced \textit{truthfulness}, \textit{informativeness}, and \textit{correctness}. However, the 70B variant diverges, excelling in \textit{truthfulness} but failing to provide useful or accurate information. We will discuss this phenomenon in the detailed LLaMA analysis. 
The Gemma series struggles with \textit{truthfulness} and \textit{correctness}, despite being \textit{informative}. This might indicate that these models are better at generating detailed content but need careful consideration for tasks requiring high accuracy or reliability.
The performance of these models highlights the complex trade-offs between being \textit{truthful}, \textit{informative}, and \textit{correct}. 
We further provide a correlation analysis between the LLM's performance and the degree/popularity of the entities in the Appendix~\ref{app:correlation_analysis}.

\paragraph{LLaMA-2 Analysis}
LLaMA-2-7B shows good \textit{truthfulness} (.965) but is moderate in being \textit{informative} (.550) and \textit{correct} (.516). This suggests that while the model is generally reliable in its outputs, it may not always provide highly detailed or accurate information.
LLaMA-2-13B significantly improves across all metrics compared to LLaMA-2-7B, with very high scores in \textit{truthfulness} (.979), \textit{informativeness} (.980), and \textit{correctness} (.959). This indicates a strong overall performance, making it a very reliable and accurate model for generating information.
LLaMA-2-70B, despite its high \textit{truthfulness} (.993), scores extremely low in both \textit{informativeness} (.007) and \textit{correctness} (.006), which is puzzling. 
We hypothesize that the model may have difficulty in making a decision, and thus selecting I don't know as the answer. This may be related to a more clear knowledge boundary of LLMs, as larger LMs tend to give up on more questions~\citep{ren2023investigating}, meaning they have a better understanding on whether they know the answer or not. This can also be confirmed by the fact that the model has the highest \textit{truthfulness} score among all models, indicating that it is more likely to provide a correct answer when it knows the answer. However, it is still important to note that a high number of `I don't know' answers may indicate the model's inability to answer factual questions. 

\paragraph{Gemma Analysis}
detailedGemma-2B has an exceptionally low \textit{truthfulness} score (.056) but is quite high in \textit{informativeness} (.867). Its \textit{correctness} score (.024) is also very low. This suggests that despite providing detailed responses, the model's outputs are often neither \textit{truthful} nor accurate. It might be generating detailed but misleading or incorrect information.
Gemma-7B improves on \textit{truthfulness} (.206) compared to Gemma-2B but still falls short of being considered reliable. Its \textit{informativeness} (.657) is respectable, and its \textit{correctness} (.056) remains low. Similar to Gemma-2B, while it can provide detailed responses, those are not often true or correct.

\begin{table}[t]
    \centering\small
    \begin{tabular}{l|ccc|ccc}
    \toprule
    \textbf{Model} & \textbf{True} & \textbf{False} & \textbf{IDK} & \textbf{Truthful} & \textbf{Informative} & \textbf{Correct} \\ 
    \midrule
    LLaMA-2-7B & 1901 & 1545 & 554 & 0.965 & 0.550 & 0.516 \\
    LLaMA-2-13B & 2100 & 1796 & 104 & 0.979 & 0.980 & 0.959 \\
    LLaMA-2-70B & 338 & 126 & 3536 & 0.993 & 0.007 & 0.006 \\
    Gemma-2B & 1760 & 1786 & 454 & 0.056 & 0.867 & 0.024 \\
    Gemma-7B & 1509 & 1751 & 740 & 0.206 & 0.657 & 0.056 \\
    \bottomrule
    \end{tabular}
    \caption{Statistics and performance metrics of LLMs. True, False, and IDK denote the number of labels from the LLMs in the labeled dataset. \Truthful, \Informative, and \Correct{} represent performance metrics.}
    \label{tab:llm_stat}
    \vspace{-3mm}
\end{table}

\vspace{-2mm}
\subsection{Relation Type Study}

\begin{figure}[t]
\centering
 \includegraphics[width=5.5in]{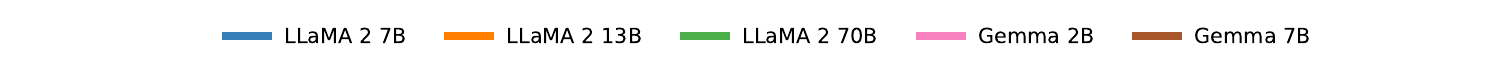}
\\\vspace{-6mm}
\subfigure[\textit{Averaged metrics} vs Head entity type]{
 \includegraphics[width=5.5in]{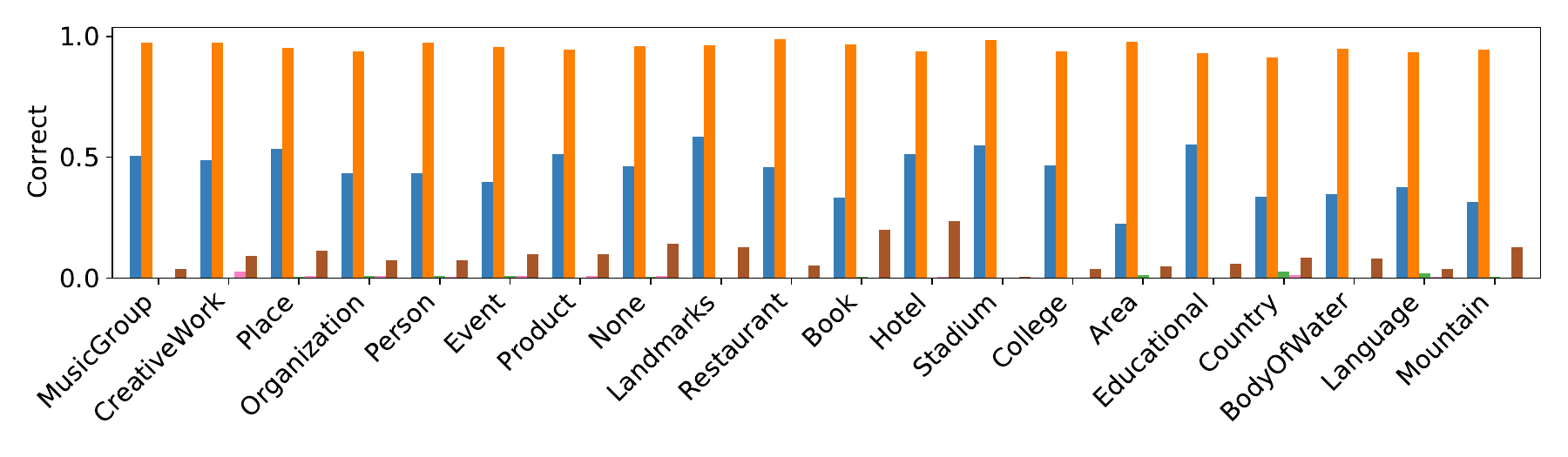}
 \label{fig:avg_by_head_type}
}\\
\vspace{-4mm}
\subfigure[\textit{Averaged metrics} vs Tail entity type]{
 \includegraphics[width=5.5in]{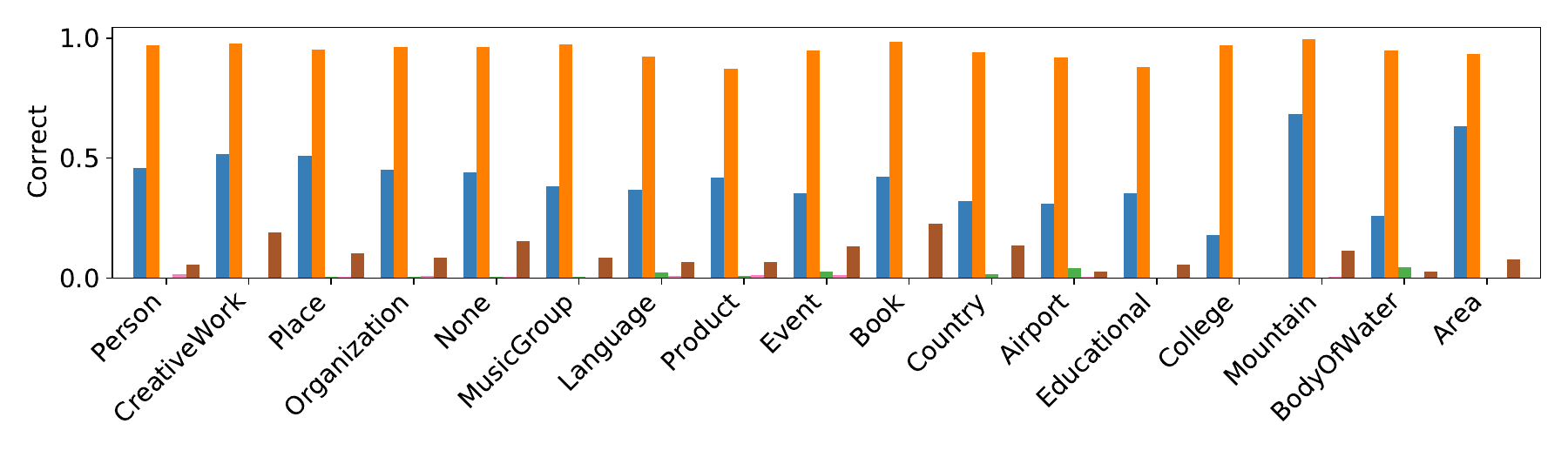}
 \label{fig:avg_by_tail_type}
}
\vspace{-4mm}
\caption{The LLM's {\it averaged metrics} with respect to head entity types and tail entity types}
\vspace{-3mm}
\label{fig:avg_by_type}
\end{figure}

\label{exp:relation_type_study} 
There are more than 600 different relation types in the DBpedia knowledge graph, and each relation type has different characteristics. It is unclear if we directly compare the performance of the LLMs on different relation types.
Thus, to gain a better understanding of the LLM's performance, we first analyze the LLM's performance with respect to relation types. In DBpedia, most entities are associated with a \url{https://schema.org/} type. Thus, we can categorize the relations into different types by the triples they belong to. We denote a relation's head/tail entity type as the most frequent schema type of the head/tail entity of the triples associated with the relation.
 For example, the relation \texttt{birthPlace} is associated with triples like \texttt{(Barack Obama, birthPlace, Hawaii)}, and the head entity \texttt{Barack Obama} is associated with the schema type \texttt{Person}, and the tail entity \texttt{Hawaii} is associated with the schema type \texttt{Place}. Then, the relation's head entity type is \texttt{Person}, and tail entity type is \texttt{Place}.
 We then analyze the LLM's performance with respect to these relation types. We report the performance of the LLMs on different relation types, by taking the average of the 3 metrics, \textit{correctness}, \textit{truthfulness}, and \textit{informativeness}, for each relation type. We present the results in Figure~\ref{fig:avg_by_type}. Here, ``None'' refers to entities not linked to a schema type. We also present a detailed analysis of the LLM's performance with respect to head and tail entity types in Appendix~\ref{app:relation_type_study}.
We can observe variability in model performance across relation types, such as ``MusicGroup" and ``CreativeWork" achieving high scores while ``Area" and "Mountain" face lower performance, highlighting the diverse challenges in modeling different kinds of information. These performance differences suggest that the effectiveness of LLMs in handling structured knowledge heavily depends on the nature of the relations being modeled.

\vspace{-0.3cm}
\section{Conclusions}
We introduce \GraphEval{}, an innovative approach for appraising the efficacy of LLMs against a voluminous test dataset derived from an extensive knowledge graph containing over 10 million facts, significantly mitigating the necessity for costly human intervention. \GraphEval{}, by embedding a judge module within the LLM itself, not only refines the evaluation process but also establishes a new benchmark for assessing the veracity of the information presented by these models. The empirical evidence from our experiments substantiates the judge model's proficiency in fact-checking, exhibiting a high degree of concordance with the accuracy of the LLM's outputs, and simultaneously diminishing the resources required for evaluation. The insights gleaned from our study shed light on the multifaceted performance of LLMs and lay the groundwork for future endeavors aimed at enhancing the reliability of their generated content.

\section*{Ethics Statement}
We use publicly available knowledge graphs and large language models and do not collect any personal data. The DBpedia knowledge graph used is retrieved from Wikipedia, in which some content, although factually correct, may be offensive to certain readers. For example, some historical events damaging to certain races/countries.  
We only evaluate the factuality of the LLMs, and do not use the LLMs for any other purposes. We hope that our work can contribute to the development of more reliable and trustworthy AI systems.

\section*{Limitations}

This research represents an initial foray into the realm of evaluating the factuality of LLMs by leveraging real-world-sized knowledge graphs, a critical step toward understanding and enhancing the reliability of LLM outputs. Amidst its pioneering efforts, several limitations have surfaced, sketching a roadmap for future investigative endeavors.
 Firstly, the judge model, while efficient in this specific context, lacks versatility for other tasks and cannot generate text, suggesting an avenue for future research to enhance its functionality and application range. Secondly, the study relies solely on zero-shot evaluation, omitting potential improvements in LLM performance through methods like in-context few-shot learning, indicating a need for more comprehensive evaluation techniques. Additionally, the white-box evaluation method employed is restricted by its requirement for access to LLM hidden states, which is not universally available, pointing to the necessity for developing black-box evaluation strategies. Lastly, the research does not examine the method's efficacy on domain-specific knowledge graphs nor consider temporal relations, marking areas for future exploration to better understand and improve LLM factuality assessments.

\bibliography{REFER} 

\begin{thebibliography}{72}
\providecommand{\natexlab}[1]{#1}
\providecommand{\url}[1]{\texttt{#1}}
\expandafter\ifx\csname urlstyle\endcsname\relax
  \providecommand{\doi}[1]{doi: #1}\else
  \providecommand{\doi}{doi: \begingroup \urlstyle{rm}\Url}\fi

\bibitem[Auer et~al.(2007)Auer, Bizer, Kobilarov, Lehmann, Cyganiak, and Ives]{auer2007dbpedia}
S{\"o}ren Auer, Christian Bizer, Georgi Kobilarov, Jens Lehmann, Richard Cyganiak, and Zachary Ives.
\newblock Dbpedia: A nucleus for a web of open data.
\newblock In \emph{International Semantic Web Conference}, pp.\  722--735. Springer, 2007.

\bibitem[Azaria \& Mitchell(2023)Azaria and Mitchell]{azaria-mitchell-2023-internal}
Amos Azaria and Tom Mitchell.
\newblock The internal state of an llm knows when it’s lying.
\newblock In \emph{Findings of the Association for Computational Linguistics: EMNLP 2023}, pp.\  967--976, 2023.

\bibitem[Ben-David et~al.(2010)Ben-David, Blitzer, Crammer, Kulesza, Pereira, and Vaughan]{ben2010theory}
Shai Ben-David, John Blitzer, Koby Crammer, Alex Kulesza, Fernando Pereira, and Jennifer~Wortman Vaughan.
\newblock A theory of learning from different domains.
\newblock \emph{Machine learning}, 79:\penalty0 151--175, 2010.

\bibitem[Berglund et~al.(2023)Berglund, Tong, Kaufmann, Balesni, Stickland, Korbak, and Evans]{berglund2023reversal}
Lukas Berglund, Meg Tong, Max Kaufmann, Mikita Balesni, Asa~Cooper Stickland, Tomasz Korbak, and Owain Evans.
\newblock The reversal curse: Llms trained on" a is b" fail to learn" b is a".
\newblock \emph{arXiv preprint arXiv:2309.12288}, 2023.

\bibitem[Bollacker et~al.(2008)Bollacker, Evans, Paritosh, Sturge, and Taylor]{bollacker2008freebase}
Kurt Bollacker, Colin Evans, Praveen Paritosh, Tim Sturge, and Jamie Taylor.
\newblock Freebase: a collaboratively created graph database for structuring human knowledge.
\newblock In \emph{Proceedings of the 2008 ACM SIGMOD international conference on Management of data}, pp.\  1247--1250, 2008.

\bibitem[Bolton et~al.(2024)Bolton, Venigalla, Yasunaga, Hall, Xiong, Lee, Daneshjou, Frankle, Liang, Carbin, and Manning]{bolton2024biomedlm}
Elliot Bolton, Abhinav Venigalla, Michihiro Yasunaga, David Hall, Betty Xiong, Tony Lee, Roxana Daneshjou, Jonathan Frankle, Percy Liang, Michael Carbin, and Christopher~D. Manning.
\newblock Biomedlm: A 2.7b parameter language model trained on biomedical text, 2024.

\bibitem[Bordes et~al.(2013)Bordes, Usunier, Garcia-Duran, Weston, and Yakhnenko]{bordes2013translating}
Antoine Bordes, Nicolas Usunier, Alberto Garcia-Duran, Jason Weston, and Oksana Yakhnenko.
\newblock Translating embeddings for modeling multi-relational data.
\newblock \emph{Advances in neural information processing systems}, 26, 2013.

\bibitem[Carlson et~al.(2010)Carlson, Betteridge, Kisiel, Settles, Hruschka, and Mitchell]{carlson2010toward}
Andrew Carlson, Justin Betteridge, Bryan Kisiel, Burr Settles, Estevam Hruschka, and Tom Mitchell.
\newblock Toward an architecture for never-ending language learning.
\newblock In \emph{Proceedings of the AAAI conference on artificial intelligence}, volume~24, pp.\  1306--1313, 2010.

\bibitem[Chang et~al.(2023)Chang, Wang, Wang, Wu, Yang, Zhu, Chen, Yi, Wang, Wang, et~al.]{EvaluationSurvey}
Yupeng Chang, Xu~Wang, Jindong Wang, Yuan Wu, Linyi Yang, Kaijie Zhu, Hao Chen, Xiaoyuan Yi, Cunxiang Wang, Yidong Wang, et~al.
\newblock A survey on evaluation of large language models.
\newblock \emph{ACM Transactions on Intelligent Systems and Technology}, 2023.

\bibitem[Chen et~al.(2016)Chen, Tian, Yang, and Zaniolo]{chen2017multilingual}
Muhao Chen, Yingtao Tian, Mohan Yang, and Carlo Zaniolo.
\newblock Multilingual knowledge graph embeddings for cross-lingual knowledge alignment.
\newblock \emph{arXiv preprint arXiv:1611.03954}, 2016.

\bibitem[Chen et~al.(2020)Chen, Hou, Cui, Che, Liu, and Yu]{chen2020recall}
Sanyuan Chen, Yutai Hou, Yiming Cui, Wanxiang Che, Ting Liu, and Xiangzhan Yu.
\newblock Recall and learn: Fine-tuning deep pretrained language models with less forgetting.
\newblock In \emph{Proceedings of the 2020 Conference on Empirical Methods in Natural Language Processing (EMNLP)}, pp.\  7870--7881, 2020.

\bibitem[Chen et~al.(2023{\natexlab{a}})Chen, Zhao, Zhang, Chern, Gao, Liu, and He]{chen2023felm}
Shiqi Chen, Yiran Zhao, Jinghan Zhang, I-Chun Chern, Siyang Gao, Pengfei Liu, and Junxian He.
\newblock {FELM}: Benchmarking factuality evaluation of large language models.
\newblock In \emph{Thirty-seventh Conference on Neural Information Processing Systems Datasets and Benchmarks Track}, 2023{\natexlab{a}}.

\bibitem[Chen et~al.(2024{\natexlab{a}})Chen, Mao, Li, Jin, Wen, Wei, Wang, Yin, Fan, Liu, and Tang]{chen2024exploring}
Zhikai Chen, Haitao Mao, Hang Li, Wei Jin, Hongzhi Wen, Xiaochi Wei, Shuaiqiang Wang, Dawei Yin, Wenqi Fan, Hui Liu, and Jiliang Tang.
\newblock Exploring the potential of large language models (llms) in learning on graphs, 2024{\natexlab{a}}.

\bibitem[Chen et~al.(2021)Chen, Chen, Geng, Pan, Yuan, and Chen]{DBLP:conf/semweb/0007CGPYC21}
Zhuo Chen, Jiaoyan Chen, Yuxia Geng, Jeff~Z Pan, Zonggang Yuan, and Huajun Chen.
\newblock Zero-shot visual question answering using knowledge graph.
\newblock In \emph{The Semantic Web--ISWC 2021: 20th International Semantic Web Conference, ISWC 2021, Virtual Event, October 24--28, 2021, Proceedings 20}, pp.\  146--162. Springer, 2021.

\bibitem[Chen et~al.(2023{\natexlab{b}})Chen, Zhang, Huang, Chen, Geng, Yu, Bi, Zhang, Yao, Song, et~al.]{DBLP:conf/icde/00070HCGYBZYSWY23}
Zhuo Chen, Wen Zhang, Yufeng Huang, Mingyang Chen, Yuxia Geng, Hongtao Yu, Zhen Bi, Yichi Zhang, Zhen Yao, Wenting Song, et~al.
\newblock Tele-knowledge pre-training for fault analysis.
\newblock In \emph{2023 IEEE 39th International Conference on Data Engineering (ICDE)}, pp.\  3453--3466. IEEE, 2023{\natexlab{b}}.

\bibitem[Chen et~al.(2024{\natexlab{b}})Chen, Zhang, Fang, Geng, Guo, Chen, Li, Zhang, Chen, Zhu, et~al.]{DBLP:journals/corr/abs-2402-05391}
Zhuo Chen, Yichi Zhang, Yin Fang, Yuxia Geng, Lingbing Guo, Xiang Chen, Qian Li, Wen Zhang, Jiaoyan Chen, Yushan Zhu, et~al.
\newblock Knowledge graphs meet multi-modal learning: A comprehensive survey.
\newblock \emph{arXiv preprint arXiv:2402.05391}, 2024{\natexlab{b}}.

\bibitem[Dao(2023)]{dao2023flashattention2}
Tri Dao.
\newblock Flashattention-2: Faster attention with better parallelism and work partitioning.
\newblock \emph{arXiv preprint arXiv:2307.08691}, 2023.

\bibitem[Diao et~al.(2023)Diao, Xu, Xu, Wang, and Zhang]{diao2023mixtureofdomainadapters}
Shizhe Diao, Tianyang Xu, Ruijia Xu, Jiawei Wang, and Tong Zhang.
\newblock Mixture-of-domain-adapters: Decoupling and injecting domain knowledge to pre-trained language models memories, 2023.

\bibitem[et~al.(2024)]{liu2024chipnemo}
Mingjie~Liu et~al.
\newblock Chipnemo: Domain-adapted llms for chip design, 2024.

\bibitem[Feng et~al.(2023)Feng, Balachandran, Bai, and Tsvetkov]{feng-etal-2023-factkb}
Shangbin Feng, Vidhisha Balachandran, Yuyang Bai, and Yulia Tsvetkov.
\newblock Factkb: Generalizable factuality evaluation using language models enhanced with factual knowledge.
\newblock In \emph{Proceedings of the 2023 Conference on Empirical Methods in Natural Language Processing}, pp.\  933--952, 2023.

\bibitem[Gallegos et~al.(2023)Gallegos, Rossi, Barrow, Tanjim, Kim, Dernoncourt, Yu, Zhang, and Ahmed]{gallegos2023bias}
Isabel~O Gallegos, Ryan~A Rossi, Joe Barrow, Md~Mehrab Tanjim, Sungchul Kim, Franck Dernoncourt, Tong Yu, Ruiyi Zhang, and Nesreen~K Ahmed.
\newblock Bias and fairness in large language models: A survey.
\newblock \emph{arXiv preprint arXiv:2309.00770}, 2023.

\bibitem[Gao et~al.(2022)Gao, Liu, Wu, Li, Wang, and Chen]{gao2022clusterea}
Yunjun Gao, Xiaoze Liu, Junyang Wu, Tianyi Li, Pengfei Wang, and Lu~Chen.
\newblock Clusterea: Scalable entity alignment with stochastic training and normalized mini-batch similarities.
\newblock In \emph{Proceedings of the 28th ACM SIGKDD Conference on Knowledge Discovery and Data Mining}, pp.\  421--431, 2022.

\bibitem[Ge et~al.(2021{\natexlab{a}})Ge, Liu, Chen, Zheng, and Gao]{ge2021largeea}
Congcong Ge, Xiaoze Liu, Lu~Chen, Baihua Zheng, and Yunjun Gao.
\newblock Largeea: Aligning entities for large-scale knowledge graphs.
\newblock \emph{Proceedings of the VLDB Endowment}, 15\penalty0 (2):\penalty0 237--245, 2021{\natexlab{a}}.

\bibitem[Ge et~al.(2021{\natexlab{b}})Ge, Liu, Chen, Zheng, and Gao]{ge2021make}
Congcong Ge, Xiaoze Liu, Lu~Chen, Baihua Zheng, and Yunjun Gao.
\newblock Make it easy: An effective end-to-end entity alignment framework.
\newblock In \emph{Proceedings of the 44th International ACM SIGIR Conference on Research and Development in Information Retrieval}, pp.\  777--786, 2021{\natexlab{b}}.

\bibitem[Goodfellow et~al.(2013)Goodfellow, Mirza, Xiao, Courville, and Bengio]{goodfellow2015empirical}
Ian~J Goodfellow, Mehdi Mirza, Da~Xiao, Aaron Courville, and Yoshua Bengio.
\newblock An empirical investigation of catastrophic forgetting in gradient-based neural networks.
\newblock \emph{arXiv preprint arXiv:1312.6211}, 2013.

\bibitem[Guo et~al.(2024)Guo, Chen, Chen, Zhang, Sun, Bo, Fang, Liu, Chen, and Zhang]{guo2024distributed}
Lingbing Guo, Zhuo Chen, Jiaoyan Chen, Yichi Zhang, Zequn Sun, Zhongpu Bo, Yin Fang, Xiaoze Liu, Huajun Chen, and Wen Zhang.
\newblock Distributed representations of entities in open-world knowledge graphs.
\newblock \emph{Knowledge-Based Systems}, pp.\  111582, 2024.

\bibitem[Hendrycks et~al.(2021)Hendrycks, Burns, Basart, Zou, Mazeika, Song, and Steinhardt]{MMLU}
Dan Hendrycks, Collin Burns, Steven Basart, Andy Zou, Mantas Mazeika, Dawn Song, and Jacob Steinhardt.
\newblock Measuring massive multitask language understanding.
\newblock \emph{Proceedings of the International Conference on Learning Representations (ICLR)}, 2021.

\bibitem[Hu et~al.(2023)Hu, Chen, Li, Guo, Wen, Yu, and Guo]{Pinocchio}
Xuming Hu, Junzhe Chen, Xiaochuan Li, Yufei Guo, Lijie Wen, Philip~S Yu, and Zhijiang Guo.
\newblock Do large language models know about facts?
\newblock \emph{arXiv preprint arXiv:2310.05177}, 2023.

\bibitem[Huang et~al.(2024)Huang, Bai, Zhu, Zhang, Zhang, Su, Liu, Lv, Zhang, Fu, et~al.]{C-Eval}
Yuzhen Huang, Yuzhuo Bai, Zhihao Zhu, Junlei Zhang, Jinghan Zhang, Tangjun Su, Junteng Liu, Chuancheng Lv, Yikai Zhang, Yao Fu, et~al.
\newblock C-eval: A multi-level multi-discipline chinese evaluation suite for foundation models.
\newblock \emph{Advances in Neural Information Processing Systems}, 36, 2024.

\bibitem[Jia et~al.(2018)Jia, Abujabal, Saha~Roy, Str{\"o}tgen, and Weikum]{TempQuestions}
Zhen Jia, Abdalghani Abujabal, Rishiraj Saha~Roy, Jannik Str{\"o}tgen, and Gerhard Weikum.
\newblock Tempquestions: A benchmark for temporal question answering.
\newblock In \emph{Companion Proceedings of the The Web Conference 2018}, pp.\  1057--1062, 2018.

\bibitem[Jiang et~al.(2023)Jiang, Zhou, Zhao, Li, and Wen]{jiang-etal-2023-reasoninglm}
Jinhao Jiang, Kun Zhou, Wayne~Xin Zhao, Yaliang Li, and Ji-Rong Wen.
\newblock Reasoninglm: Enabling structural subgraph reasoning in pre-trained language models for question answering over knowledge graph.
\newblock In \emph{Proceedings of the 2023 Conference on Empirical Methods in Natural Language Processing}, pp.\  3721--3735, 2023.

\bibitem[Joshi et~al.(2017)Joshi, Choi, Weld, and Zettlemoyer]{TQ}
Mandar Joshi, Eunsol Choi, Daniel~S Weld, and Luke Zettlemoyer.
\newblock Triviaqa: A large scale distantly supervised challenge dataset for reading comprehension.
\newblock In \emph{Proceedings of the 55th Annual Meeting of the Association for Computational Linguistics (Volume 1: Long Papers)}, pp.\  1601--1611, 2017.

\bibitem[Kasai et~al.(2024)Kasai, Sakaguchi, Le~Bras, Asai, Yu, Radev, Smith, Choi, Inui, et~al.]{kasai2022realtimeqa}
Jungo Kasai, Keisuke Sakaguchi, Ronan Le~Bras, Akari Asai, Xinyan Yu, Dragomir Radev, Noah~A Smith, Yejin Choi, Kentaro Inui, et~al.
\newblock Realtime qa: What's the answer right now?
\newblock \emph{Advances in Neural Information Processing Systems}, 36, 2024.

\bibitem[Kim et~al.(2023)Kim, Kwon, Jo, and Choi]{kim2023kggpt}
Jiho Kim, Yeonsu Kwon, Yohan Jo, and Edward Choi.
\newblock Kg-gpt: A general framework for reasoning on knowledge graphs using large language models.
\newblock \emph{arXiv preprint arXiv:2310.11220}, 2023.

\bibitem[Kotha et~al.(2023)Kotha, Springer, and Raghunathan]{kotha2023understanding}
Suhas Kotha, Jacob~Mitchell Springer, and Aditi Raghunathan.
\newblock Understanding catastrophic forgetting in language models via implicit inference.
\newblock \emph{arXiv preprint arXiv:2309.10105}, 2023.

\bibitem[Kwiatkowski et~al.(2019)Kwiatkowski, Palomaki, Redfield, Collins, Parikh, Alberti, Epstein, Polosukhin, Devlin, Lee, et~al.]{NaturalQuestions}
Tom Kwiatkowski, Jennimaria Palomaki, Olivia Redfield, Michael Collins, Ankur Parikh, Chris Alberti, Danielle Epstein, Illia Polosukhin, Jacob Devlin, Kenton Lee, et~al.
\newblock Natural questions: A benchmark for question answering research.
\newblock \emph{Transactions of the Association for Computational Linguistics}, 7:\penalty0 453--466, 2019.

\bibitem[Lewis et~al.(2020)Lewis, Perez, Piktus, Petroni, Karpukhin, Goyal, K{\"u}ttler, Lewis, Yih, Rockt{\"a}schel, et~al.]{lewis2020retrieval}
Patrick Lewis, Ethan Perez, Aleksandra Piktus, Fabio Petroni, Vladimir Karpukhin, Naman Goyal, Heinrich K{\"u}ttler, Mike Lewis, Wen-tau Yih, Tim Rockt{\"a}schel, et~al.
\newblock Retrieval-augmented generation for knowledge-intensive nlp tasks.
\newblock \emph{Advances in Neural Information Processing Systems}, 33:\penalty0 9459--9474, 2020.

\bibitem[Li et~al.(2023)Li, Cheng, Zhao, Nie, and Wen]{HaluEval}
Junyi Li, Xiaoxue Cheng, Xin Zhao, Jian-Yun Nie, and Ji-Rong Wen.
\newblock Halueval: A large-scale hallucination evaluation benchmark for large language models.
\newblock In \emph{The 2023 Conference on Empirical Methods in Natural Language Processing}, 2023.

\bibitem[Liang et~al.(2022)Liang, Bommasani, Lee, Tsipras, Soylu, Yasunaga, Zhang, Narayanan, Wu, Kumar, et~al.]{liang2023holistic}
Percy Liang, Rishi Bommasani, Tony Lee, Dimitris Tsipras, Dilara Soylu, Michihiro Yasunaga, Yian Zhang, Deepak Narayanan, Yuhuai Wu, Ananya Kumar, et~al.
\newblock Holistic evaluation of language models.
\newblock \emph{arXiv preprint arXiv:2211.09110}, 2022.

\bibitem[Lin et~al.(2022)Lin, Hilton, and Evans]{TruthfulQA}
Stephanie Lin, Jacob Hilton, and Owain Evans.
\newblock Truthfulqa: Measuring how models mimic human falsehoods.
\newblock In \emph{Proceedings of the 60th Annual Meeting of the Association for Computational Linguistics (Volume 1: Long Papers)}, pp.\  3214--3252, 2022.

\bibitem[Liu et~al.(2023{\natexlab{a}})Liu, Wu, Michael, Suhr, West, Koller, Swayamdipta, Smith, and Choi]{liu2023we}
Alisa Liu, Zhaofeng Wu, Julian Michael, Alane Suhr, Peter West, Alexander Koller, Swabha Swayamdipta, Noah~A Smith, and Yejin Choi.
\newblock We're afraid language models aren't modeling ambiguity.
\newblock \emph{arXiv preprint arXiv:2304.14399}, 2023{\natexlab{a}}.

\bibitem[Liu et~al.(2021)Liu, Ji, Fu, Tam, Du, Yang, and Tang]{liu2021p}
Xiao Liu, Kaixuan Ji, Yicheng Fu, Weng~Lam Tam, Zhengxiao Du, Zhilin Yang, and Jie Tang.
\newblock P-tuning v2: Prompt tuning can be comparable to fine-tuning universally across scales and tasks.
\newblock \emph{arXiv preprint arXiv:2110.07602}, 2021.

\bibitem[Liu et~al.(2022)Liu, Ji, Fu, Tam, Du, Yang, and Tang]{liu-etal-2022-p}
Xiao Liu, Kaixuan Ji, Yicheng Fu, Weng Tam, Zhengxiao Du, Zhilin Yang, and Jie Tang.
\newblock P-tuning: Prompt tuning can be comparable to fine-tuning across scales and tasks.
\newblock In \emph{Proceedings of the 60th Annual Meeting of the Association for Computational Linguistics (Volume 2: Short Papers)}, pp.\  61--68, 2022.

\bibitem[Liu et~al.(2023{\natexlab{b}})Liu, Wu, Li, Chen, and Gao]{liu2023unsupervised}
Xiaoze Liu, Junyang Wu, Tianyi Li, Lu~Chen, and Yunjun Gao.
\newblock Unsupervised entity alignment for temporal knowledge graphs.
\newblock In \emph{Proceedings of the ACM Web Conference 2023}, pp.\  2528--2538, 2023{\natexlab{b}}.

\bibitem[Lu et~al.(2022)Lu, Mishra, Xia, Qiu, Chang, Zhu, Tafjord, Clark, and Kalyan]{ScienceQA}
Pan Lu, Swaroop Mishra, Tony Xia, Liang Qiu, Kai-Wei Chang, Song-Chun Zhu, Oyvind Tafjord, Peter Clark, and Ashwin Kalyan.
\newblock Learn to explain: Multimodal reasoning via thought chains for science question answering.
\newblock In \emph{Advances in Neural Information Processing Systems}, 2022.

\bibitem[Luo et~al.(2023)Luo, Li, Haffari, and Pan]{luo2023reasoning}
Linhao Luo, Yuan-Fang Li, Gholamreza Haffari, and Shirui Pan.
\newblock Reasoning on graphs: Faithful and interpretable large language model reasoning.
\newblock \emph{arXiv preprint arXiv:2310.01061}, 2023.

\bibitem[Ren et~al.(2023)Ren, Wang, Qu, Zhao, Liu, Tian, Wu, Wen, and Wang]{ren2023investigating}
Ruiyang Ren, Yuhao Wang, Yingqi Qu, Wayne~Xin Zhao, Jing Liu, Hao Tian, Hua Wu, Ji-Rong Wen, and Haifeng Wang.
\newblock Investigating the factual knowledge boundary of large language models with retrieval augmentation.
\newblock \emph{arXiv preprint arXiv:2307.11019}, 2023.

\bibitem[Srivastava et~al.(2023)Srivastava, Rastogi, Rao, Shoeb, Abid, Fisch, Brown, Santoro, Gupta, Garriga-Alonso, et~al.]{BigBench}
Aarohi Srivastava, Abhinav Rastogi, Abhishek Rao, Abu Awal~Md Shoeb, Abubakar Abid, Adam Fisch, Adam~R Brown, Adam Santoro, Aditya Gupta, Adri{\`a} Garriga-Alonso, et~al.
\newblock Beyond the imitation game: Quantifying and extrapolating the capabilities of language models.
\newblock \emph{Transactions on Machine Learning Research}, 2023.

\bibitem[Suchanek et~al.(2007)Suchanek, Kasneci, and Weikum]{suchanek2007yago}
Fabian~M Suchanek, Gjergji Kasneci, and Gerhard Weikum.
\newblock Yago: a core of semantic knowledge.
\newblock In \emph{Proceedings of the 16th international conference on World Wide Web}, pp.\  697--706, 2007.

\bibitem[Sun et~al.(2023)Sun, Xu, Zha, Liu, and Dong]{sun2023head}
Kai Sun, Yifan~Ethan Xu, Hanwen Zha, Yue Liu, and Xin~Luna Dong.
\newblock Head-to-tail: How knowledgeable are large language models (llm)? aka will llms replace knowledge graphs?
\newblock \emph{arXiv preprint arXiv:2308.10168}, 2023.

\bibitem[Sun et~al.(2019)Sun, Deng, Nie, and Tang]{sun2019rotate}
Zhiqing Sun, Zhi-Hong Deng, Jian-Yun Nie, and Jian Tang.
\newblock Rotate: Knowledge graph embedding by relational rotation in complex space.
\newblock \emph{arXiv preprint arXiv:1902.10197}, 2019.

\bibitem[Tan et~al.(2023)Tan, Min, Li, Li, Hu, Chen, and Qi]{tan2023chatgpt}
Yiming Tan, Dehai Min, Yu~Li, Wenbo Li, Nan Hu, Yongrui Chen, and Guilin Qi.
\newblock Can chatgpt replace traditional kbqa models? an in-depth analysis of the question answering performance of the gpt llm family.
\newblock In \emph{International Semantic Web Conference}, pp.\  348--367. Springer, 2023.

\bibitem[Tan et~al.(2021)Tan, Cai, Dong, and Ma]{9499743}
Zhanhong Tan, Hongyu Cai, Runpei Dong, and Kaisheng Ma.
\newblock Nn-baton: Dnn workload orchestration and chiplet granularity exploration for multichip accelerators.
\newblock In \emph{2021 ACM/IEEE 48th Annual International Symposium on Computer Architecture (ISCA)}, pp.\  1013--1026, 2021.

\bibitem[Team et~al.(2024)Team, Mesnard, Hardin, Dadashi, Bhupatiraju, Pathak, Sifre, Rivi{\`e}re, Kale, Love, et~al.]{gemmateam2024gemma}
Gemma Team, Thomas Mesnard, Cassidy Hardin, Robert Dadashi, Surya Bhupatiraju, Shreya Pathak, Laurent Sifre, Morgane Rivi{\`e}re, Mihir~Sanjay Kale, Juliette Love, et~al.
\newblock Gemma: Open models based on gemini research and technology.
\newblock \emph{arXiv preprint arXiv:2403.08295}, 2024.

\bibitem[Tian et~al.(2023)Tian, Mitchell, Yao, Manning, and Finn]{tian2023finetuning}
Katherine Tian, Eric Mitchell, Huaxiu Yao, Christopher~D Manning, and Chelsea Finn.
\newblock Fine-tuning language models for factuality.
\newblock \emph{arXiv preprint arXiv:2311.08401}, 2023.

\bibitem[Touvron et~al.(2023)Touvron, Martin, Stone, Albert, Almahairi, Babaei, Bashlykov, Batra, Bhargava, Bhosale, et~al.]{touvron2023llama}
Hugo Touvron, Louis Martin, Kevin Stone, Peter Albert, Amjad Almahairi, Yasmine Babaei, Nikolay Bashlykov, Soumya Batra, Prajjwal Bhargava, Shruti Bhosale, et~al.
\newblock Llama 2: Open foundation and fine-tuned chat models.
\newblock \emph{arXiv preprint arXiv:2307.09288}, 2023.

\bibitem[Vu et~al.(2023)Vu, Iyyer, Wang, Constant, Wei, Wei, Tar, Sung, Zhou, Le, et~al.]{vu2023freshllms}
Tu~Vu, Mohit Iyyer, Xuezhi Wang, Noah Constant, Jerry Wei, Jason Wei, Chris Tar, Yun-Hsuan Sung, Denny Zhou, Quoc Le, et~al.
\newblock Freshllms: Refreshing large language models with search engine augmentation.
\newblock \emph{arXiv preprint arXiv:2310.03214}, 2023.

\bibitem[Wang et~al.(2023{\natexlab{a}})Wang, Cheng, Guo, Yue, Ding, Xu, Wang, Hu, Zhang, and Zhang]{wang2023evaluating}
Cunxiang Wang, Sirui Cheng, Qipeng Guo, Yuanhao Yue, Bowen Ding, Zhikun Xu, Yidong Wang, Xiangkun Hu, Zheng Zhang, and Yue Zhang.
\newblock Evaluating open-{QA} evaluation.
\newblock In \emph{Thirty-seventh Conference on Neural Information Processing Systems Datasets and Benchmarks Track}, 2023{\natexlab{a}}.

\bibitem[Wang et~al.(2023{\natexlab{b}})Wang, Liu, Yue, Tang, Zhang, Jiayang, Yao, Gao, Hu, Qi, et~al.]{wang2023survey}
Cunxiang Wang, Xiaoze Liu, Yuanhao Yue, Xiangru Tang, Tianhang Zhang, Cheng Jiayang, Yunzhi Yao, Wenyang Gao, Xuming Hu, Zehan Qi, et~al.
\newblock Survey on factuality in large language models: Knowledge, retrieval and domain-specificity.
\newblock \emph{arXiv preprint arXiv:2310.07521}, 2023{\natexlab{b}}.

\bibitem[Wang et~al.(2024{\natexlab{a}})Wang, Zhao, and Gao]{wang2024blendfilter}
Haoyu Wang, Tuo Zhao, and Jing Gao.
\newblock Blendfilter: Advancing retrieval-augmented large language models via query generation blending and knowledge filtering, 2024{\natexlab{a}}.

\bibitem[Wang et~al.(2024{\natexlab{b}})Wang, Ma, Hu, Weber-Genzel, R{\"o}ttger, Kreuter, Hovy, and Plank]{wang2024myanswer}
Xinpeng Wang, Bolei Ma, Chengzhi Hu, Leon Weber-Genzel, Paul R{\"o}ttger, Frauke Kreuter, Dirk Hovy, and Barbara Plank.
\newblock " my answer is c": First-token probabilities do not match text answers in instruction-tuned language models.
\newblock \emph{arXiv preprint arXiv:2402.14499}, 2024{\natexlab{b}}.

\bibitem[Wang et~al.(2022)Wang, Si, Li, Lukasik, Yu, Hsieh, Dhillon, and Kumar]{wang2022preserving}
Yihan Wang, Si~Si, Daliang Li, Michal Lukasik, Felix Yu, Cho-Jui Hsieh, Inderjit~S Dhillon, and Sanjiv Kumar.
\newblock Preserving in-context learning ability in large language model fine-tuning.
\newblock \emph{arXiv preprint arXiv:2211.00635}, 2022.

\bibitem[Weller et~al.(2023)Weller, Marone, Weir, Lawrie, Khashabi, and Van~Durme]{weller2023according}
Orion Weller, Marc Marone, Nathaniel Weir, Dawn Lawrie, Daniel Khashabi, and Benjamin Van~Durme.
\newblock " according to..." prompting language models improves quoting from pre-training data.
\newblock \emph{arXiv preprint arXiv:2305.13252}, 2023.

\bibitem[Yao et~al.(2023)Yao, Wang, Tian, Cheng, Li, Deng, Chen, and Zhang]{yao2023editing}
Yunzhi Yao, Peng Wang, Bozhong Tian, Siyuan Cheng, Zhoubo Li, Shumin Deng, Huajun Chen, and Ningyu Zhang.
\newblock Editing large language models: Problems, methods, and opportunities.
\newblock \emph{arXiv preprint arXiv:2305.13172}, 2023.

\bibitem[Yasunaga et~al.(2022)Yasunaga, Bosselut, Ren, Zhang, Manning, Liang, and Leskovec]{yasunaga2022deep}
Michihiro Yasunaga, Antoine Bosselut, Hongyu Ren, Xikun Zhang, Christopher~D Manning, Percy~S Liang, and Jure Leskovec.
\newblock Deep bidirectional language-knowledge graph pretraining.
\newblock volume~35, pp.\  37309--37323, 2022.

\bibitem[Yin et~al.(2023)Yin, Sun, Guo, Wu, Qiu, and Huang]{yin-etal-2023-large}
Zhangyue Yin, Qiushi Sun, Qipeng Guo, Jiawen Wu, Xipeng Qiu, and Xuan-Jing Huang.
\newblock Do large language models know what they don’t know?
\newblock In \emph{Findings of the Association for Computational Linguistics: ACL 2023}, pp.\  8653--8665, 2023.

\bibitem[Zhai et~al.(2023)Zhai, Tong, Li, Cai, Qu, Lee, and Ma]{zhai2023investigating}
Yuexiang Zhai, Shengbang Tong, Xiao Li, Mu~Cai, Qing Qu, Yong~Jae Lee, and Yi~Ma.
\newblock Investigating the catastrophic forgetting in multimodal large language models.
\newblock \emph{arXiv preprint arXiv:2309.10313}, 2023.

\bibitem[Zhang et~al.(2024)Zhang, Ye, Liu, Ren, Wu, and Chen]{zhang2024knowledge}
Mengqi Zhang, Xiaotian Ye, Qiang Liu, Pengjie Ren, Shu Wu, and Zhumin Chen.
\newblock Knowledge graph enhanced large language model editing.
\newblock \emph{arXiv preprint arXiv:2402.13593}, 2024.

\bibitem[Zhang et~al.(2023{\natexlab{a}})Zhang, Chen, Fang, Cheng, Lu, Li, Zhang, and Chen]{DBLP:journals/corr/abs-2311-06503}
Yichi Zhang, Zhuo Chen, Yin Fang, Lei Cheng, Yanxi Lu, Fangming Li, Wen Zhang, and Huajun Chen.
\newblock Knowledgeable preference alignment for llms in domain-specific question answering.
\newblock \emph{arXiv preprint arXiv:2311.06503}, 2023{\natexlab{a}}.

\bibitem[Zhang et~al.(2023{\natexlab{b}})Zhang, Chen, Zhang, and Chen]{DBLP:journals/corr/abs-2310-06671}
Yichi Zhang, Zhuo Chen, Wen Zhang, and Huajun Chen.
\newblock Making large language models perform better in knowledge graph completion.
\newblock \emph{arXiv preprint arXiv:2310.06671}, 2023{\natexlab{b}}.

\bibitem[Zhang et~al.(2023{\natexlab{c}})Zhang, Li, Cui, Cai, Liu, Fu, Huang, Zhao, Zhang, Chen, et~al.]{zhang2023siren}
Yue Zhang, Yafu Li, Leyang Cui, Deng Cai, Lemao Liu, Tingchen Fu, Xinting Huang, Enbo Zhao, Yu~Zhang, Yulong Chen, et~al.
\newblock Siren's song in the ai ocean: a survey on hallucination in large language models.
\newblock \emph{arXiv preprint arXiv:2309.01219}, 2023{\natexlab{c}}.

\bibitem[Zhou et~al.(2023)Zhou, Zhu, Chen, Chen, Zhao, Chen, Lin, Wen, and Han]{zhou2023dont}
Kun Zhou, Yutao Zhu, Zhipeng Chen, Wentong Chen, Wayne~Xin Zhao, Xu~Chen, Yankai Lin, Ji-Rong Wen, and Jiawei Han.
\newblock Don't make your llm an evaluation benchmark cheater.
\newblock \emph{arXiv preprint arXiv:2311.01964}, 2023.

\end{thebibliography}
\bibliographystyle{colm2024_conference}

\appendix

\newpage
\section{Appendix}

\subsection{Judge Model Analysis}
\label{app:judge_model}
\begin{figure}[t]
\captionsetup[subfigure]{aboveskip=1pt,belowskip=1pt}
    \centering
    \includegraphics[width=5.5in]{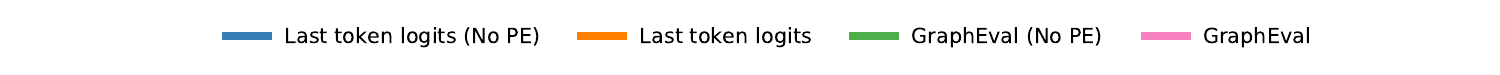}\\
     \vspace{-5mm}
    \subfigure[LLaMA 7B]{
        \includegraphics[width=1.78in]{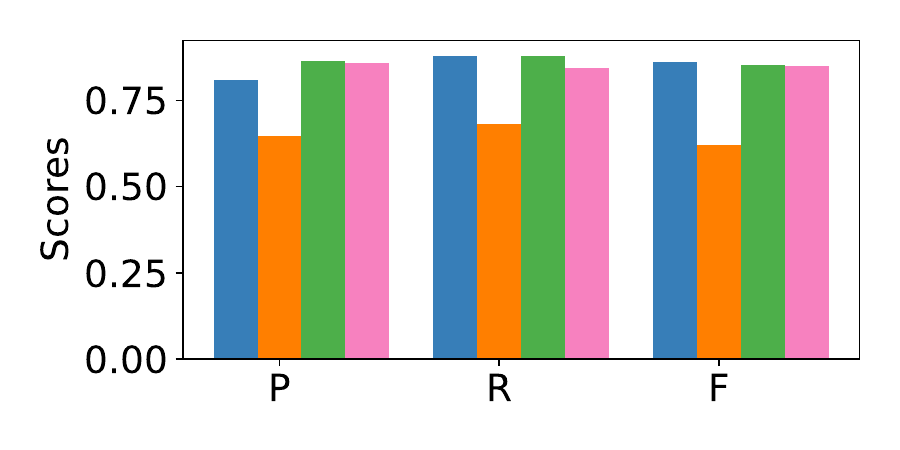}
        \label{fig:evaluation_scores_llama_2_7b}
       }\hspace{-4mm}
    \subfigure[LLaMA 13B]{
        \includegraphics[width=1.78in]{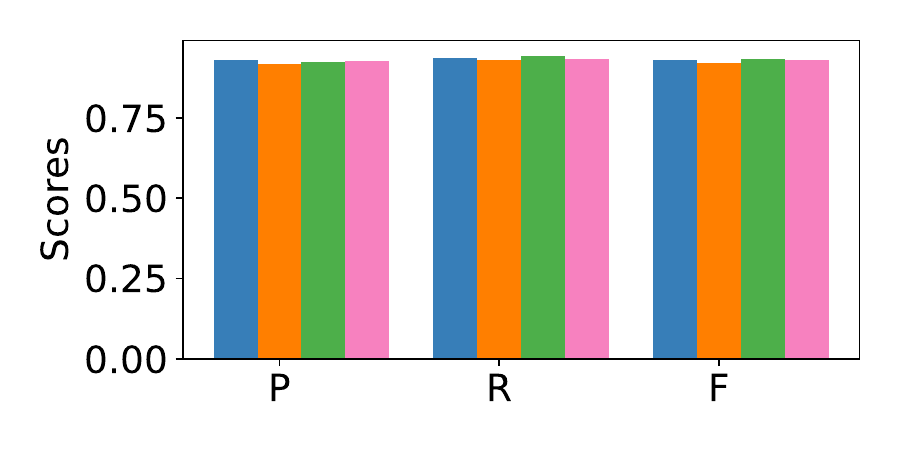}
        \label{fig:evaluation_scores_llama_2_13b}
       }\hspace{-4mm}
    \subfigure[LLaMA 70B]{
        \includegraphics[width=1.78in]{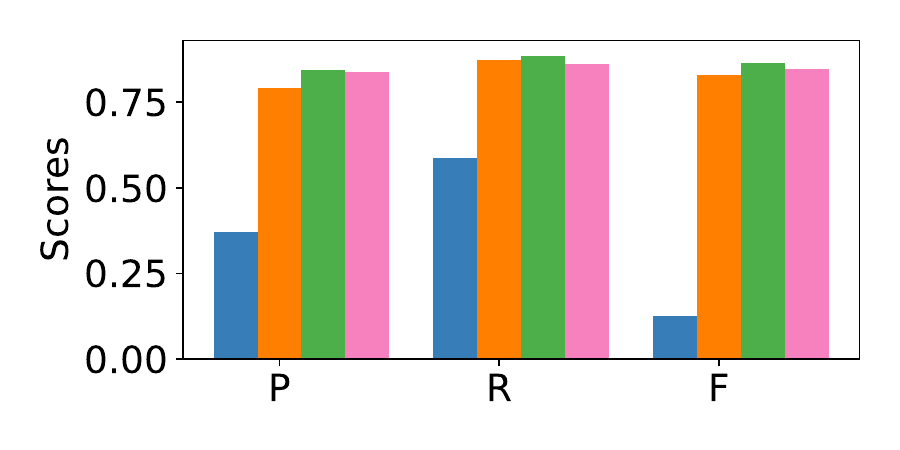}
        \label{fig:evaluation_scores_llama_2_70b}
       }
    \\\vspace{-4mm}
    \subfigure[Gemma 2B]{
        \includegraphics[width=1.78in]{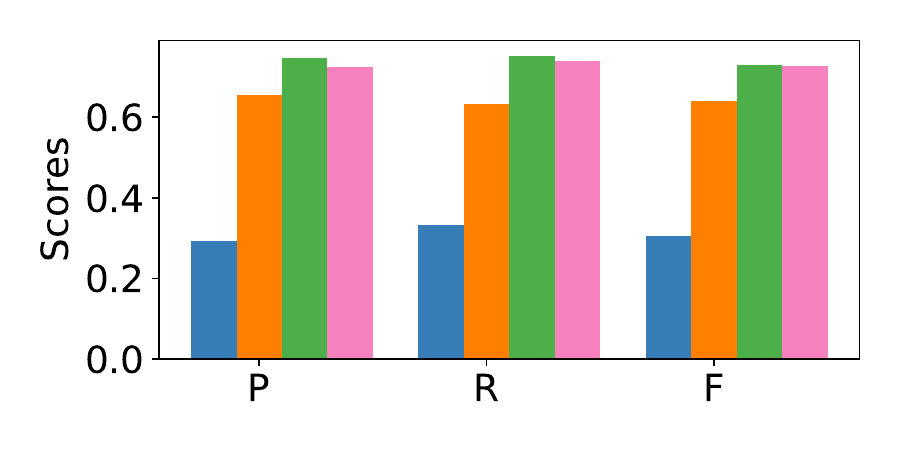}
        \label{fig:evaluation_scores_gemma_2b}
       }\hspace{-4mm}
    \subfigure[Gemma 7B]{
        \includegraphics[width=1.78in]{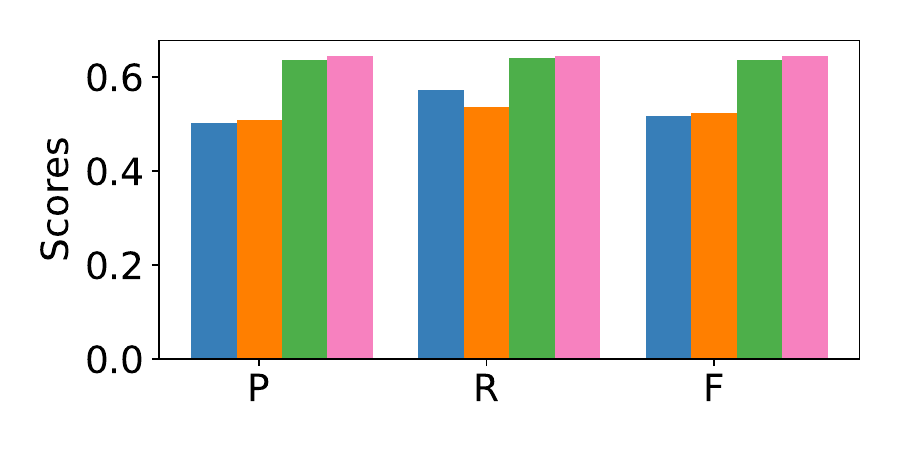}
        \label{fig:evaluation_scores_gemma_7b}
       }\hspace{-4mm}
    \subfigure[Average]{
        \includegraphics[width=1.78in]{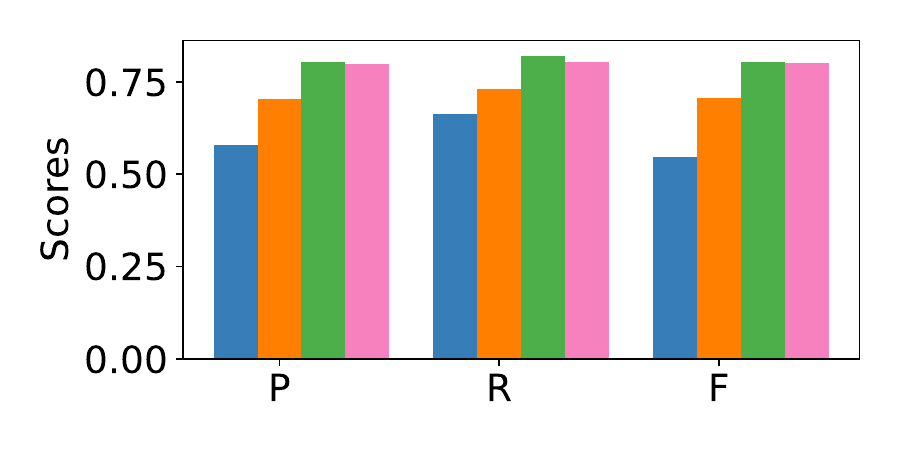}
        \label{fig:evaluation_scores_average}
       }
       \vspace{-3mm}
    \caption{Evaluation scores on the judge model's performance on the labeled validation set. P, R, and F are Precision, Recall, and F1 Score. 
    }
    \vspace{-3mm}
    \label{fig:evaluation_scores}
\end{figure}

We analyze the judge model's performance on the labeled validation set. 
We compare \GraphEval{}'s judge model by using the last token logit as the judge model. This is a common practice in evaluating LLMs, as the last token logit is the most common way to extract the hidden state of the LLMs. We also analyze the judge model with or without the prompt encoder (PE), as it may have a negative impact on the judge model's performance. We refer to Figure~\ref{fig:evaluation_scores} for the judge model's performance on the labeled validation set.

\begin{table}[t]
    \centering
    \setlength{\tabcolsep}{2mm}{
    \begin{tabular}{lcccccccccc}
    \toprule
    \multirow{2}{*}{\textbf{Substitute Model}}  & \multicolumn{3}{c}{\textbf{LLaMA 2 7B}} & \multicolumn{3}{c}{\textbf{LLaMA 2 13B}} & \multicolumn{3}{c}{\textbf{LLaMA 2 70B}} \\
    \cmidrule(lr){2-4} \cmidrule(lr){5-7} \cmidrule(lr){8-10} 
    
    & P & R & F & P & R & F & P & R & F  \\
    \midrule
    \textbf{LLaMA 2 7B} & .858 & .845 & .850 & .928 & .934 & .930 & .837 & .861 & .848 \\
    \textbf{LLaMA 2 13B} & .850 & .868 & .855 & .930 & .940 & .932 & .837 & .851 & .844 \\
    \textbf{LLaMA 2 70B} & .868 & .883 & .871 & .924 & .942 & .931 & .858 & .876 & .866 \\
 
    \bottomrule
    \end{tabular}}
    \caption{Ablation on the LLaMA models as substitute models. The $i$-th row and $j$-th column denote the result of using $i$-th LLM as the substitute hidden state input for training on $j$-th model's labels.  P, R, and F are Precision, Recall, and F1 Score.  
    }
    \label{tab:abla_substitute}
    \end{table}

\begin{table}[t]
    \centering
    \small
    \setlength{\tabcolsep}{.5mm}{
    \begin{tabular}{lcccccccccc}
    \toprule
    \multirow{2}{*}{\textbf{Models}}
    & \multicolumn{2}{c}{\textbf{LLaMA 2 7B}} & \multicolumn{2}{c}{\textbf{LLaMA 2 13B}} & \multicolumn{2}{c}{\textbf{LLaMA 2 70B}} & \multicolumn{2}{c}{\textbf{Gemma 2B}} & \multicolumn{2}{c}{\textbf{Gemma7B}} \\
    \cmidrule(lr){2-3} \cmidrule(lr){4-5} \cmidrule(lr){6-7} \cmidrule(lr){8-9} \cmidrule(lr){10-11}
      & Speed & \#GPUs   & Speed & \#GPUs  & Speed & \#GPUs & Speed & \#GPUs & Speed & \#GPUs \\
    \midrule
    TG (A6000) & 2.26 & 1 & 1.07 & 2 & 0.09 & 4 & 2.06 (1.82) & 1 & 2.18 (1.28) & 1 \\
    \GraphEval{} (A6000) & 121.34  & 1 & 120.10 & 1 & 117.90 & 1 & 388.61 & 1 & 389.04 & 1 \\
    \midrule
    TG (A100) &  2.80 & 1 & 1.48 & 1 & 0.21 & 2 & 2.47 & 1 & 2.42 & 1 \\
    \GraphEval{} (A100) &  210.59 & 1 & 213.05 & 1 & 210.30 & 1 & 731.98 & 1 & 735.62 & 1 \\
    \bottomrule
    \end{tabular}}
    \caption{Efficiency evaluation. Speed denotes the average number of triple facts on which a conclusion can be given in one second. \#GPUs denotes the least number of GPUs to run without OOM. TG denotes text generation. 
     The numbers in parentheses are the speed without Flash Attention 2.}
    \label{tab:speed_test}
\end{table}

 \paragraph{Accuracy Analysis}
 The  \GraphEval{} model, both with and without Prompt Encoder (PE), consistently outperforms the score of using Last token logits in almost all configurations and metrics. This indicates the effectiveness of the  \GraphEval{} approach in capturing the nuances of the evaluation task.

 \paragraph{Ablation Study}
{\it On Prompt Encoder:}
As Figure~\ref{fig:evaluation_scores} shows, 
the comparison between models with and without PE indicates a slight performance variation. For \GraphEval{}, the presence of PE does not significantly alter the performance, suggesting that our method of evaluating LLMs is robust to the inclusion or exclusion of PE.
For the Last token logits method, removing PE generally results in a perturbation in performance. However, the  \GraphEval{} approach's consistency suggests a potentially different or more advanced mechanism of evaluation that is less dependent on PE.
{\it On Substitute Models:}
We also evaluate the judge model's performance on different LLMs as hidden state input. 
We refer to Table \ref{tab:abla_substitute} for the judge model's performance on different LLMs as hidden state input. We can see that, generally, when larger models are applied for feeding the hidden states, there is a slight increase in the fitting accuracy of the judge model. However, there is no significant difference in the judge model's performance.

\paragraph{Efficiency study}
We also analyze the judge model's efficiency by measuring the time it takes to make a prediction on one triple.
The speed of text generation refers to the average rate at which the LLM completes generating a response consisting of one sentence derived from a triple. It's important to recognize that the pace of text generation can vary with different prompts because the LLM may produce responses of varying lengths. Therefore, for a more consistent measure of text generation speeds, it's advisable to consider the rate of token generation. Despite this, our evaluation framework, \GraphEval{}, does not depend on text generation and operates on a triple-based unit. Consequently, we continue to use the triple as the unit of measurement for time.
We use the same hardware and software environment for all the experiments. We compare the average speed of the judge model with text generation. We report the time it takes to make a prediction in Table \ref{tab:speed_test}. 
 The attention implementation and precision are the same for text generation and for the judge model's input model. 
 We can see that the judge model is significantly faster than text generation. This indicates that the judge model is efficient in evaluating the LLMs. Also, benefiting from the substitute model, our evaluation speed and GPU requirement does not grow with the LLM size, which is an advantage for evaluating large LLMs.
 We also observe that, paradoxically, the Gemma 2B model operates slower than the Gemma 7B model, despite its smaller size. This counterintuitive result could be attributed to the implementation of Flash Attention 2. To draw a fair comparison, we documented the text generation speed on A6000 GPUs excluding Flash Attention 2, which is indicated within parentheses. The comparative data reveals that Gemma 2B is faster than Gemma 7B when Flash Attention 2 is not utilized. Notwithstanding this, Gemma 2B demonstrates enhanced performance when Flash Attention 2 is active. Therefore, for the sake of consistency, we have decided to maintain the results acquired with Flash Attention 2.

\subsection{Detailed Relation Type Analysis}
\label{app:relation_type_study}

\begin{figure}[t]
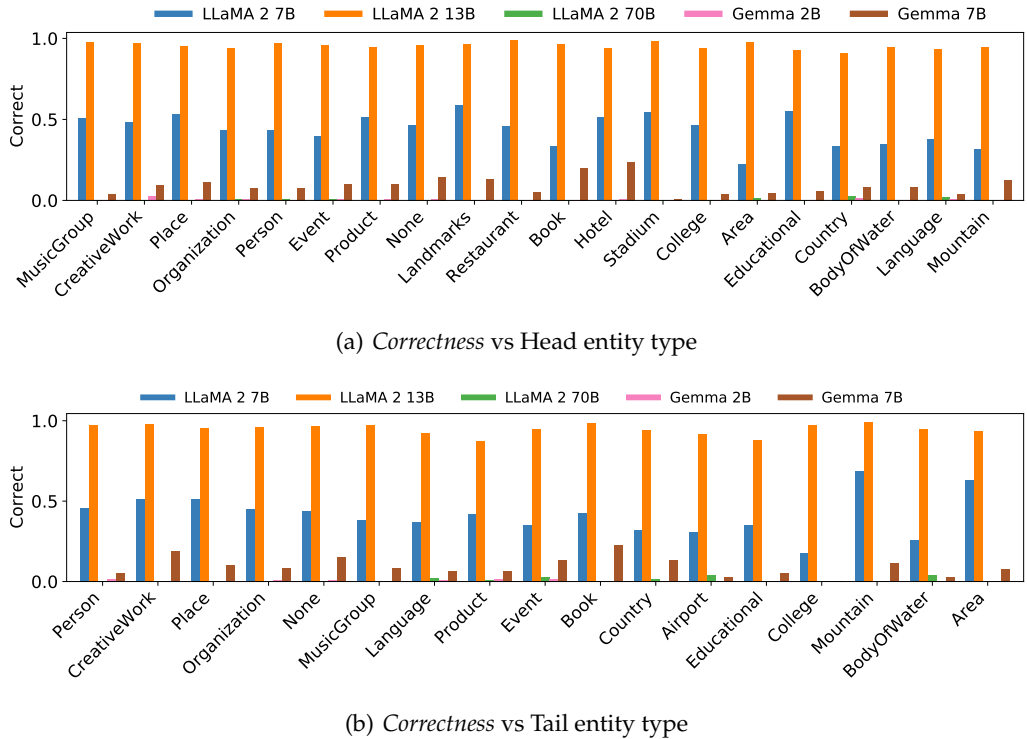

\centering
 \includegraphics[width=6in]{figures/experiments/relation_analysis/legend.pdf}
\\\vspace{-6mm}
\subfigure[\textit{Correctness} vs Head entity type]{
 \includegraphics[width=5.5in]{figures/experiments/relation_analysis/correct_by_head_type.pdf}
 \label{fig:correct_by_head_type}
}\\
\includegraphics[width=5.5in]{figures/experiments/relation_analysis/legend.pdf}
\\\vspace{-6mm}
\subfigure[\textit{Correctness} vs Tail entity type]{
 \includegraphics[width=5.5in]{figures/experiments/relation_analysis/correct_by_tail_type.pdf}
 \label{fig:correct_by_tail_type}
}
\caption{The LLM's \textit{correctness} with respect to head entity types and tail entity types}
\label{fig:correctness_by_type}
\end{figure}

\begin{figure}[t]
    \centering
 \includegraphics[width=5.5in]{figures/experiments/relation_analysis/legend.pdf}
 \\\vspace{-6mm}
\subfigure[\textit{Truthfulness} vs Head entity type]{
 \includegraphics[width=5.5in]{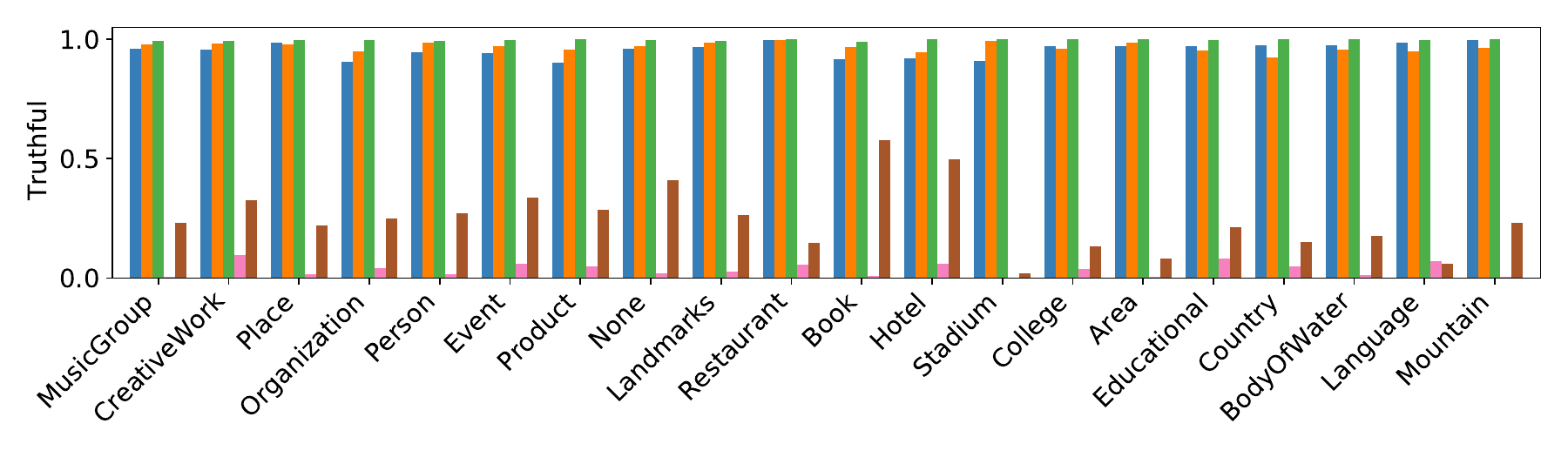}
 \label{fig:truthful_by_head_type}
}\\
\includegraphics[width=5.5in]{figures/experiments/relation_analysis/legend.pdf}
\\\vspace{-6mm}
\subfigure[\textit{Truthfulness} vs Tail entity type]{
 \includegraphics[width=5.5in]{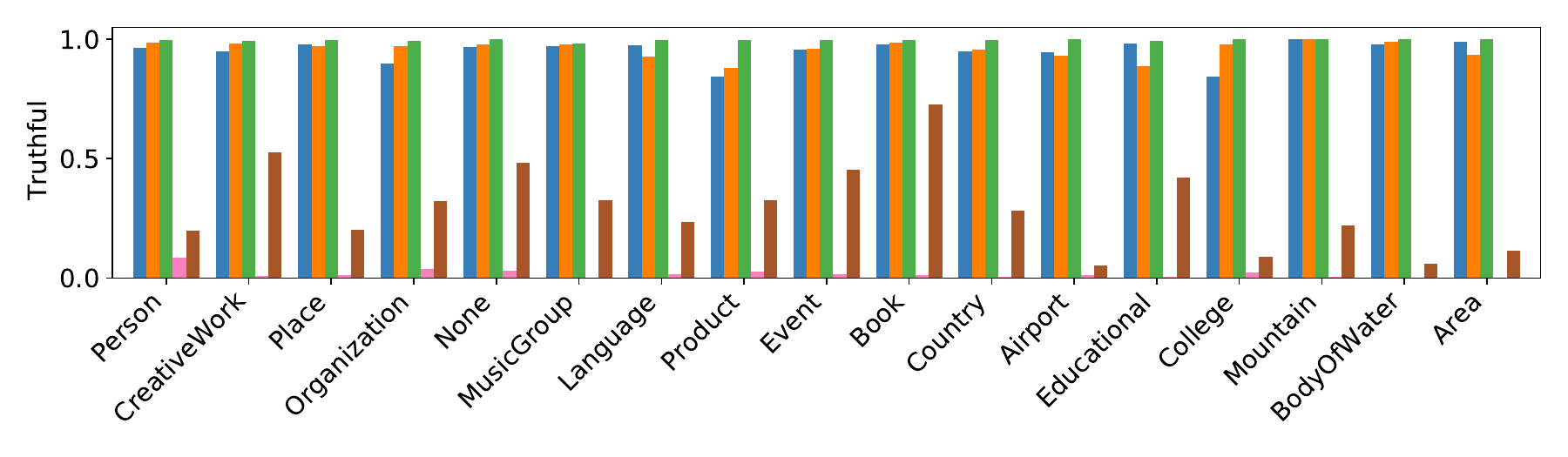}
 \label{fig:truthful_by_tail_type}
}
\caption{The LLM's \textit{truthfulness} with respect to head entity types and tail entity types}
\label{fig:truthfulness_by_type}
\end{figure}
\begin{figure}[t]
    \centering
    \includegraphics[width=6in]{figures/experiments/relation_analysis/legend.pdf}
    \\\vspace{-6mm}
\subfigure[\textit{Informativeness} vs Head entity type]{
 \includegraphics[width=5.5in]{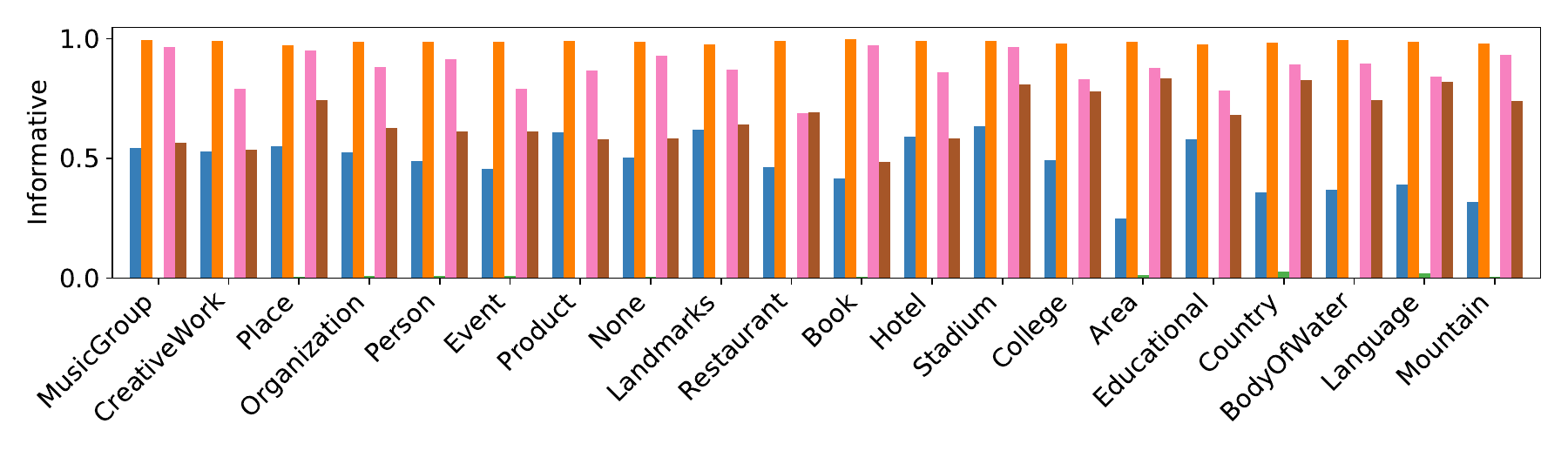}
 \label{fig:informative_by_head_type}
}\\
\includegraphics[width=5.5in]{figures/experiments/relation_analysis/legend.pdf}
\\\vspace{-6mm}
\subfigure[\textit{Informativeness} vs Tail entity type]{
 \includegraphics[width=5.5in]{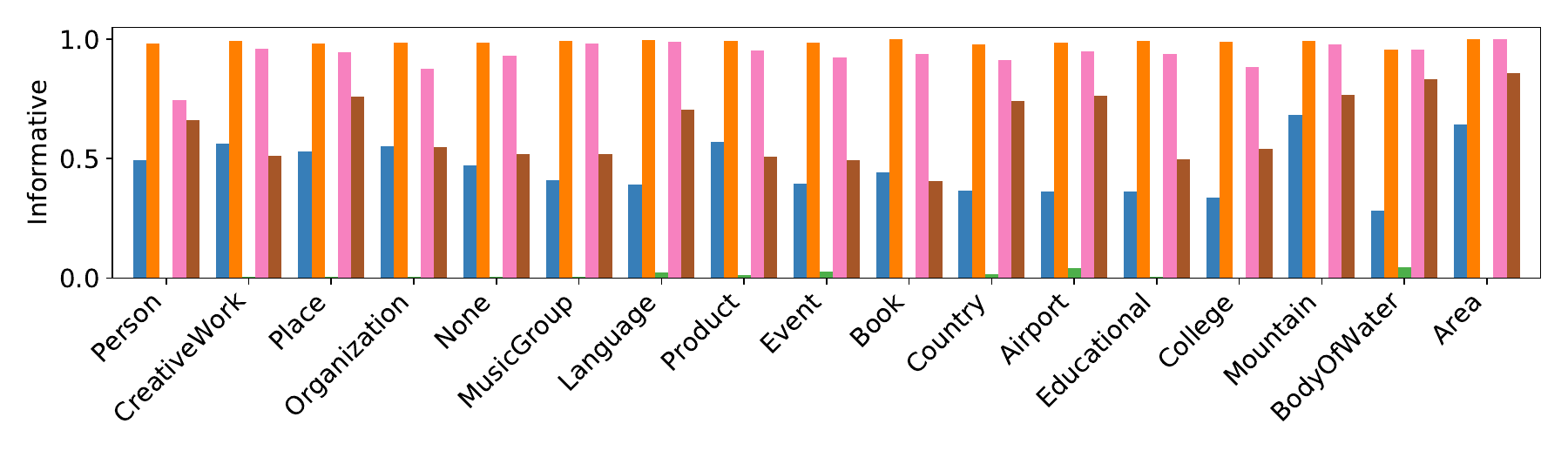}
 \label{fig:informative_by_tail_type}
}
\caption{The LLM's \textit{informativeness} with respect to head entity types and tail entity types}
\label{fig:informativeness_by_type}
\end{figure}

\paragraph{Llama Family Analysis}
Across the LLaMA family, a progressive improvement in performance is observed from  7b to 13b. The 7b model shows decent performance across categories with a particular strength in the \textit{truthfulness}. However, its \textit{informativeness} and \textit{correctness} metrics show room for improvement, particularly in categories like Book, Hotel, and College, indicating a struggle to accurately provide informative and correct classifications in more nuanced or specific domains.

The LLaMA 13b model demonstrates a significant leap in performance, especially in \textit{informativeness} and \textit{correctness}, nearly reaching perfection across most categories. This jump can be attributed to the model's increased capacity, enabling it to understand and process the nuances of various entities better, resulting in remarkably high scores in nearly all categories, especially noticeable in MusicGroup, CreativeWork, and Place.

The LLaMA 70b results appear anomalous with extremely high \textit{truthfulness} scores but negligible \textit{informativeness} and \textit{correctness} across all categories. We suspect this discrepancy might be due to the model's knowledge awareness~\cite{ren2023investigating}, where the model might be less confident in its responses when the parameters are increased, leading to a higher proportion of ``I don't know" responses. This could explain the high \textit{truthfulness} scores but low \textit{informativeness} and \textit{correctness} metrics, as the model might be too cautious to provide definitive answers.

\paragraph{Gemma Family Analysis}
The Gemma models present an interesting contrast. The Gemma 2b model shows a tendency towards high \textit{informativeness} in certain categories like MusicGroup and Book but lacks behind significantly in \textit{truthfulness} and \textit{correctness} metrics. This suggests that while the model might be picking up on relevant information, it struggles to accurately validate the truth behind that information or its applicability to the queried entities.
The Gemma 7b model shows improvement in the \textit{truthfulness} metric compared to Gemma 2b, particularly noticeable in categories like Book and Hotel, and even surpasses LLaMA 7b in certain areas like None and Restaurant. However, it still significantly lags behind the LLaMA models, particularly LLaMA 13b, in both \textit{informativeness} and \textit{correctness}. The improved but still limited performance suggests that while Gemma 7b has a better grasp over the veracity of information compared to Gemma 2b, it still struggles with providing highly informative and correct outputs consistently across various entities.

\subsection{Correlation Analysis}
\label{app:correlation_analysis}
As the current language models are all exposed to Wikipedia knowledge during training, we are interested in how the LLM performance is correlated with the attributes of the triples in the knowledge graphs. As an example, if an entity has a higher degree, it may be linked to more documents, and the LLM may have more chances to learn about the entity during training. Another example is the popularity of the entity. If the entity is more popular, it may be linked to more external documents, and the LLM may have more chances to learn about the entity during training. This raises the question of whether the LLM's performance is correlated with the attributes of the triples in the knowledge graphs.
For the entities in a knowledge graph, the degree of an entity is the number of edges connected to the entity. We also collect the \textit{pageviews} of the entities in the knowledge graph from Wikimedia\footnote{https://wikimedia.org/api/rest\_v1/metrics/pageviews/per-article/en.wikipedia.org/all-access/all-agents}{}, which is the number of pageviews of the Wikipedia page of the entity. This can be seen as a measure of the popularity of the entity. We collect the pageviews, in the time period of the entities in the knowledge graph from the Wikipedia page of the entity.
After collecting the degree and pageviews of the entities in the knowledge graph, we can aggregate the degree and pageviews of the entities to the triples, by simply taking the average of the degree and pageviews of the head and tail entities of the triples.

\begin{figure}[t]
\centering
\subfigure[LLaMA-2-7B]{
 \includegraphics[width=1.6in]{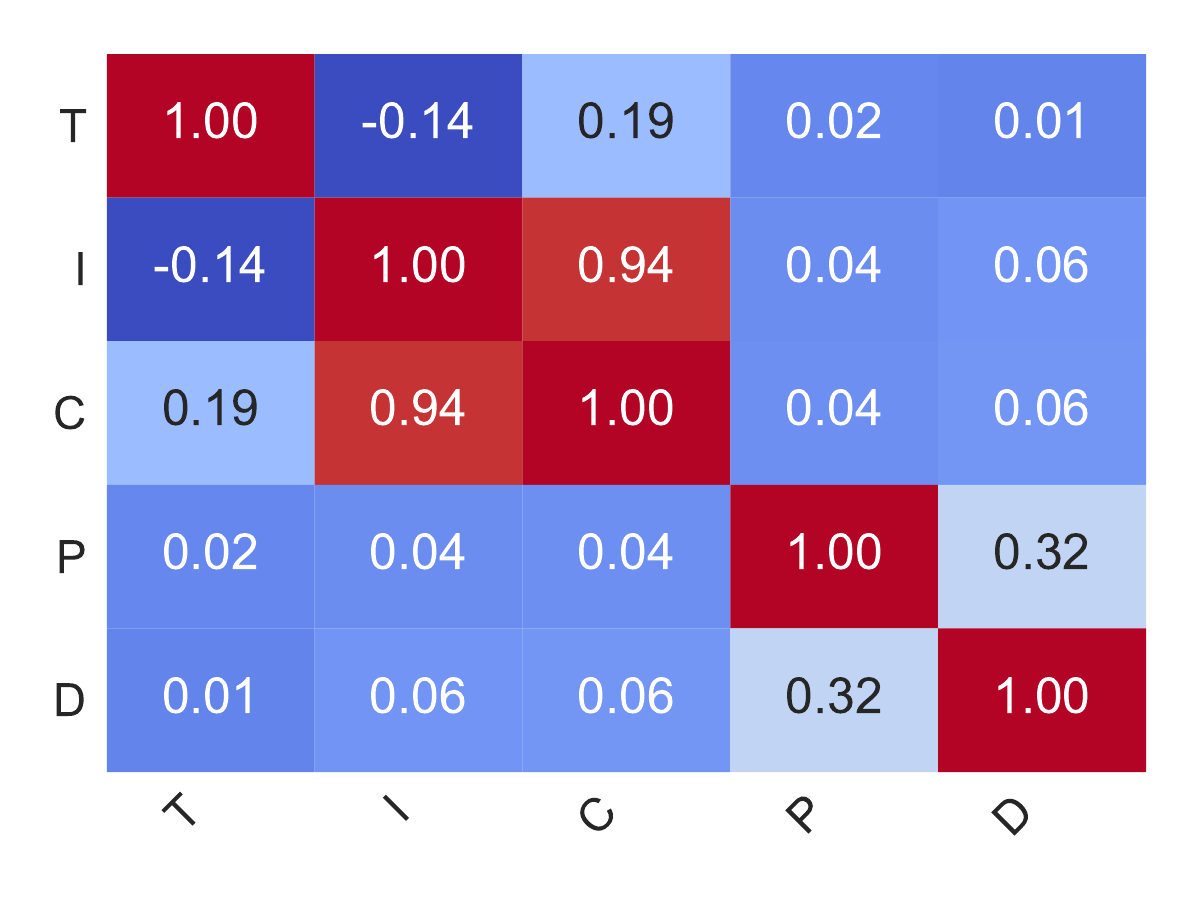}
 \label{fig:correlation_heatmap_llama_7b}
} 
\subfigure[LLaMA-2-13B]{
 \includegraphics[width=1.6in]{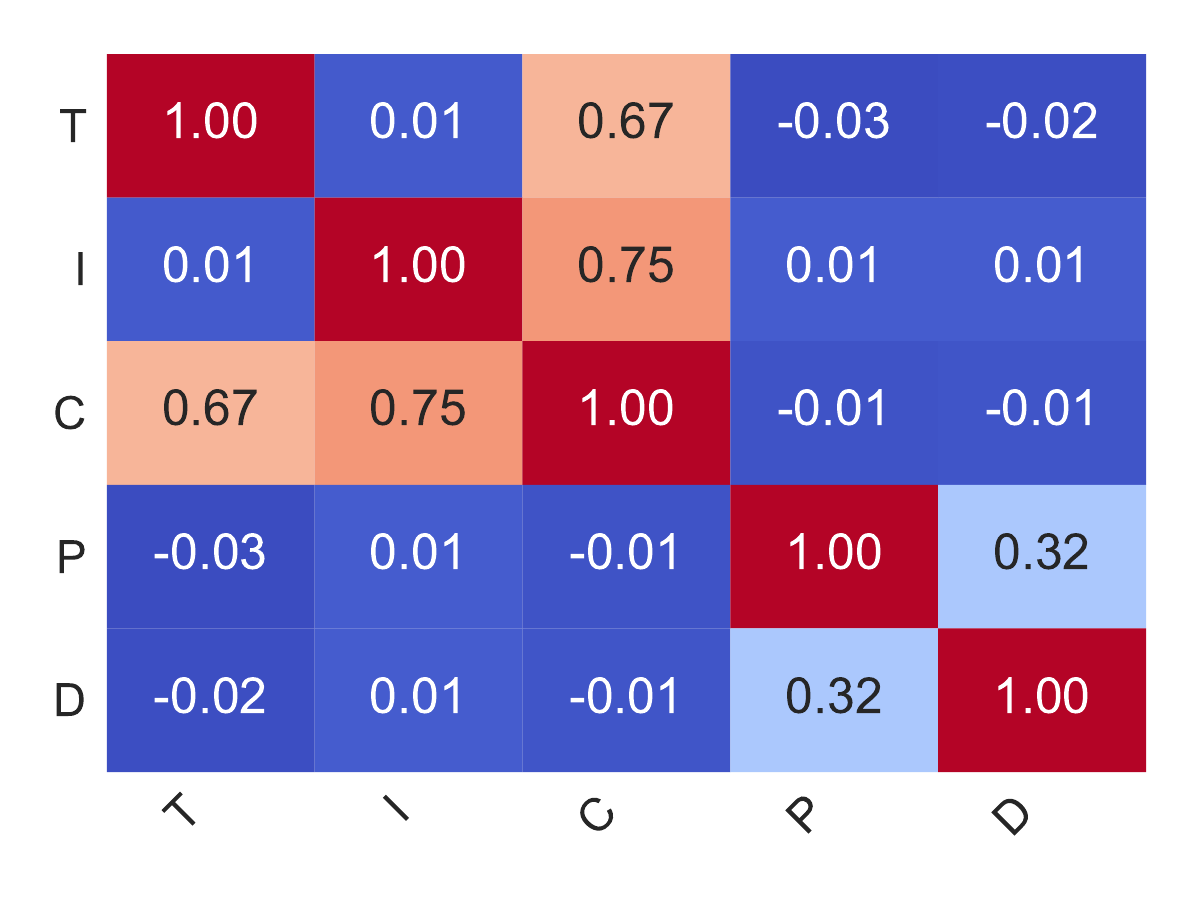}
 \label{fig:correlation_heatmap_llama_13b}
} 
\subfigure[LLaMA-2-70B]{
 \includegraphics[width=1.6in]{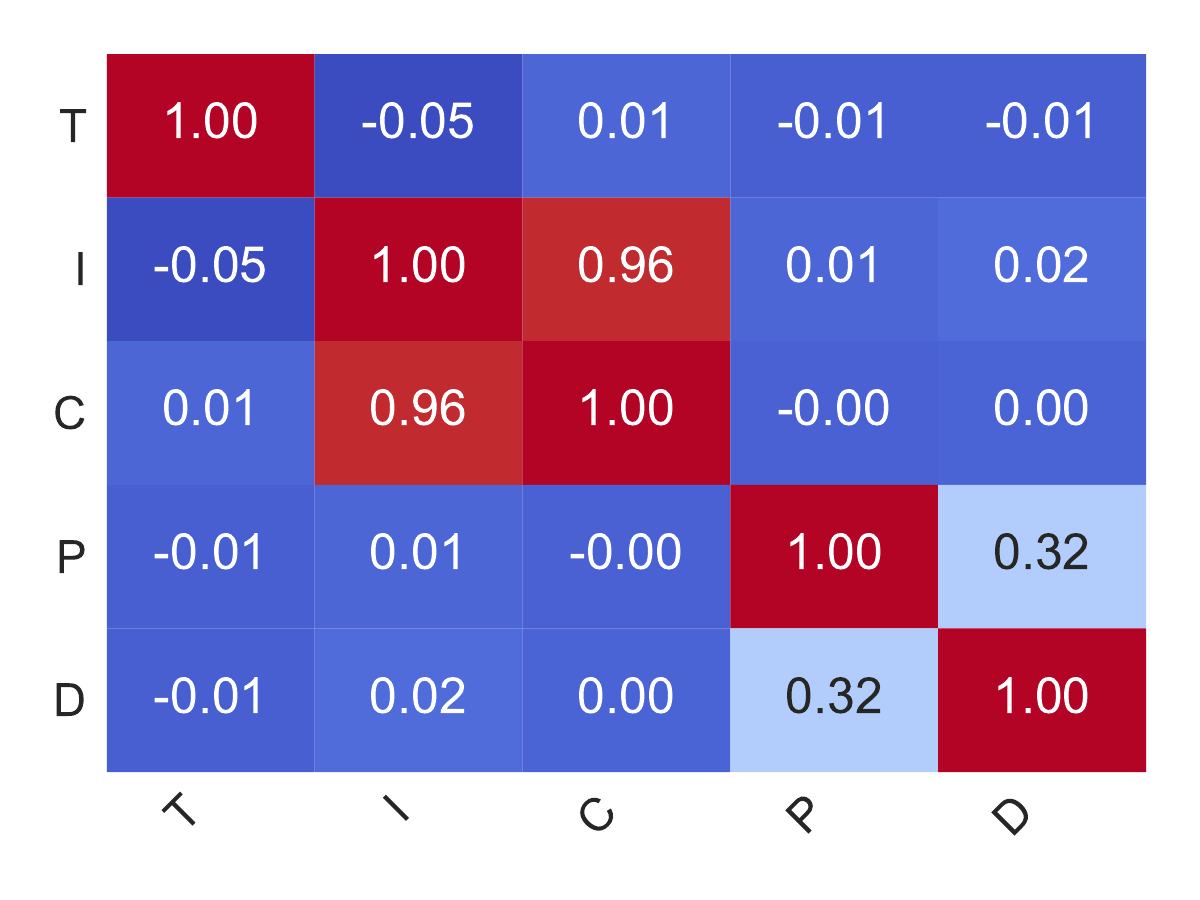}
 \label{fig:correlation_heatmap_llama_70b}
}
\\
\subfigure[Gemma-2B]{
 \includegraphics[width=1.7in]{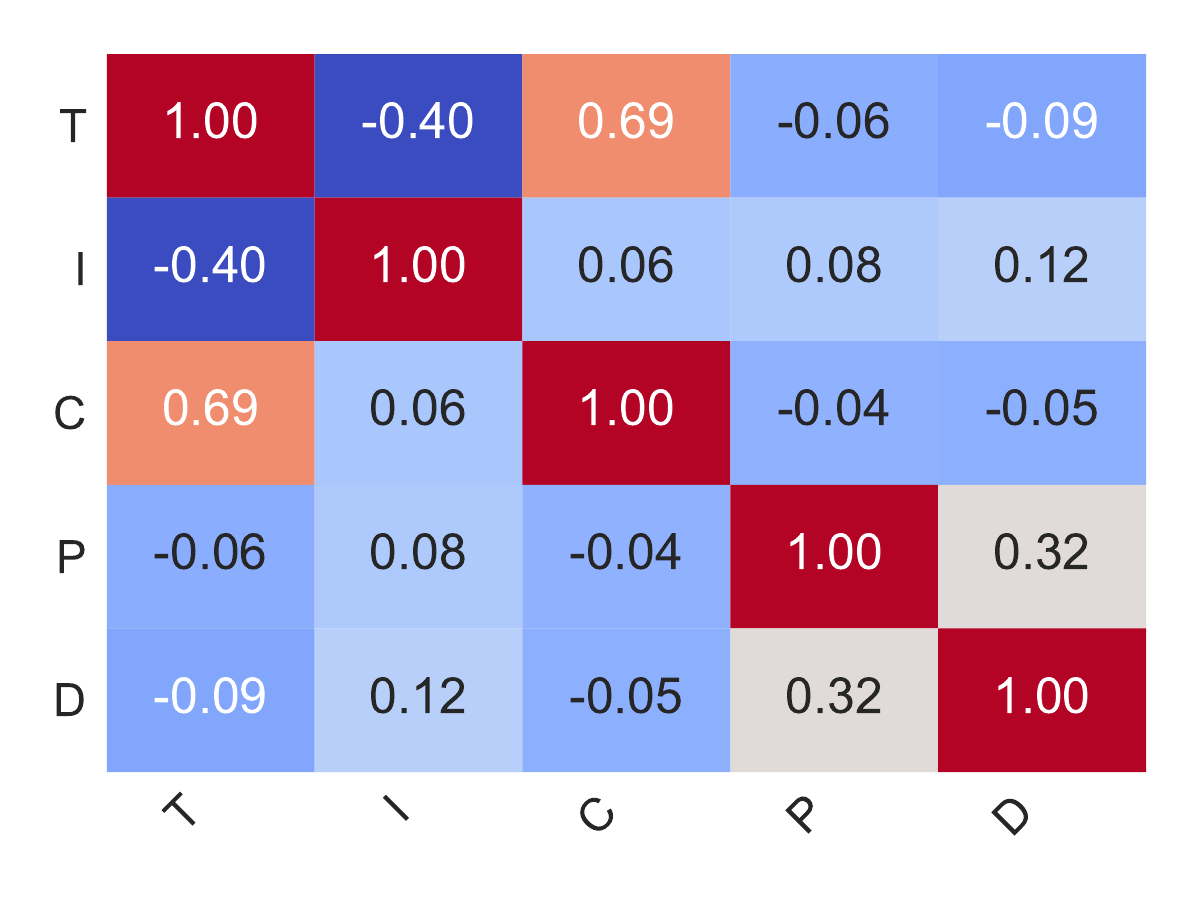}
 \label{fig:correlation_heatmap_gemma_2b}
}
\subfigure[Gemma-7B]{
 \includegraphics[width=1.7in]{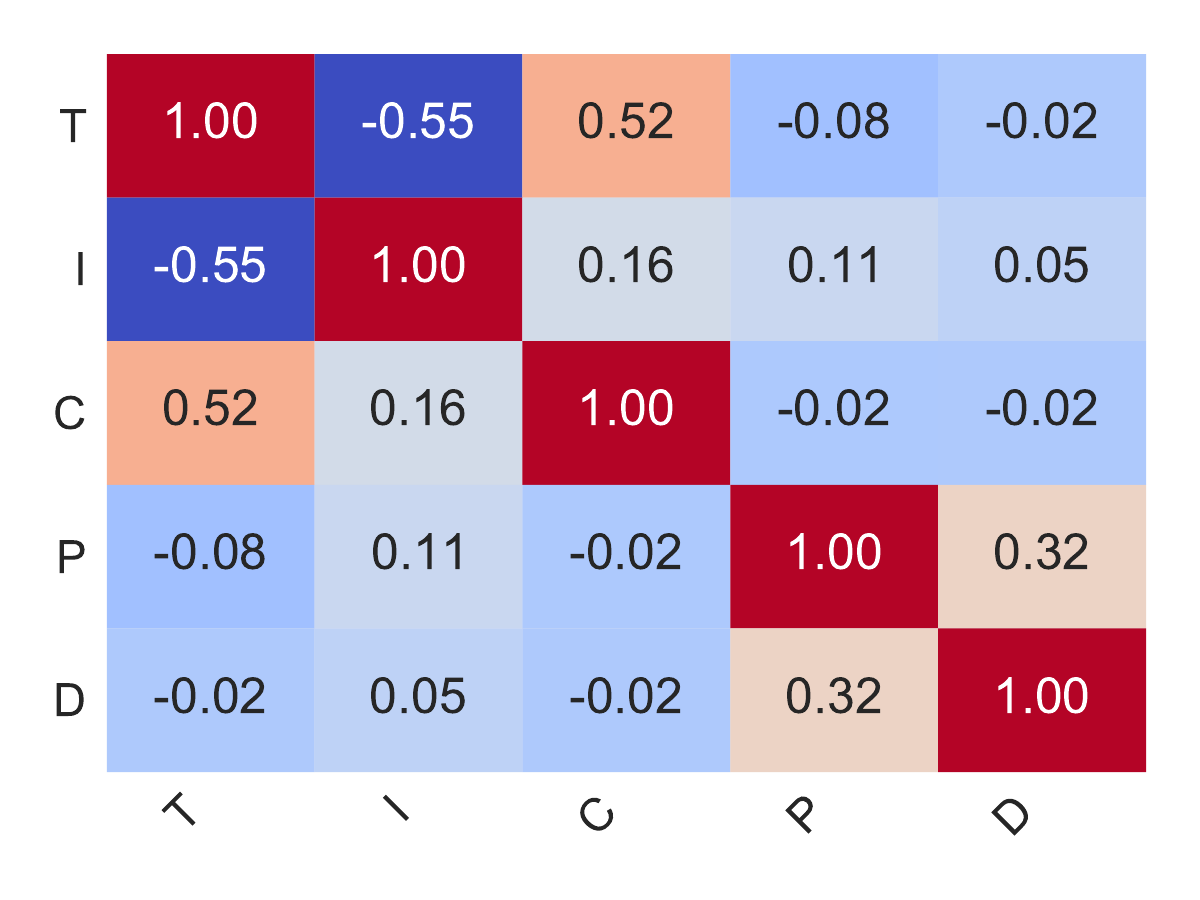}
 \label{fig:correlation_heatmap_gemma_7b}
}
\caption{Correlation heatmap of the LLMs' hidden states and the judge model's predictions.}
\label{fig:correlation_heatmap}
\end{figure}

Here, we analyze whether the LLM's performance is correlated with the attributes of the triples in the knowledge graphs, such as the entity's degree, and page views. 
We refer to Figure \ref{fig:correlation_heatmap} for the correlation heatmap of the LLMs' hidden states and the judge model's predictions. Here, `T' stands for \textit{Truthful}, `I' stands for \textit{Informativeness}, `C' stands for \textit{Correctness}, `P' stands for Pageviews, and `D' stands for Degree. 
 We can see that the LLM's performance does not show a strong correlation with the attributes of the triples in the knowledge graphs. This indicates that the LLM's performance is not directly correlated with the attributes of the triples in the knowledge graphs. However, the different metrics of LLMs may correlate with each other, such as \textit{Truthful} and \textit{Informativeness}, which is expected. 

\subsection{Detailed Settings}
\label{app:detailed_settings}

\begin{figure}[t]
    \centering
    \includegraphics[width=4.5in]{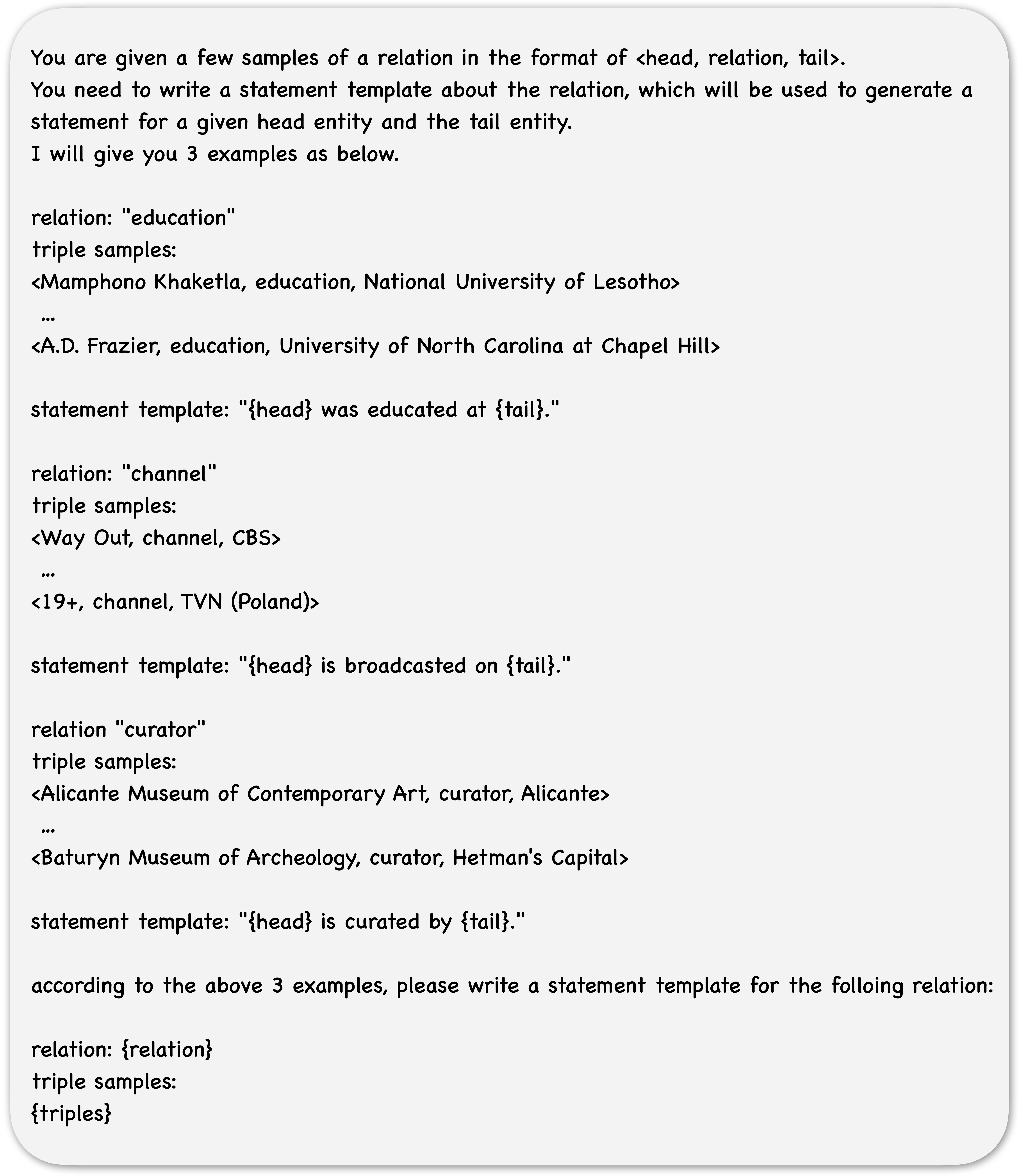}
    \vspace{-3mm}
    \caption{The prompt to generate the relation template.}
    \label{fig:relation_template_prompt}
\end{figure}

\begin{table}[t]
    \centering\small\setlength{\tabcolsep}{0.2in}{
    \begin{tabular}{l|l|l}
    \toprule
    \textbf{Relation} & \textbf{Template} & \textbf{Count} \\
    \midrule

        birthPlace & The birthplace of \{head\} is \{tail\}. & 1,465,157 \\
    team & \{head\} is a part of the \{tail\} team. & 1,265,483 \\
    subdivision & The subdivision of \{head\} is \{tail\}. & 1,070,387 \\
    country &  \{head\} is from the country \{tail\}. & 766,844 \\
    starring &  \{head\} is a character in a movie or play \{tail\}". & 540,937 \\
    location & The location of \{head\} is \{tail\}. & 523,283 \\
    type & The type of \{head\} is \{tail\}. & 480,274 \\
    deathPlace & \{head\} passed away in \{tail\}. & 435,869 \\
    timeZone & The time zone of \{head\} is \{tail\}. & 433,915 \\
    genre & The genre of \{head\} is \{tail\}. & 415,336 \\
    homepage & The homepage of \{head\} is \{tail\}. & 366,745 \\
    position & The position of \{head\} is \{tail\}. & 319,196 \\
    seeAlso & The related item to \{head\} under the  & 296,615 \\
    & $\;\hookrightarrow$label 'seeAlso' is \{tail\}. & \\
    writer & The writer of \{head\} is \{tail\}. & 249,017 \\
    almaMater & The alma mater of \{head\} is \{tail\}. & 217,533 \\
    occupation & The occupation of \{head\} is \{tail\}. & 200,615 \\
    award & The award won by \{head\} is \{tail\}. & 181,521 \\
    recordLabel & The record label associated with \{head\} is \{tail\}. & 178,657 \\
    party & The party that \{head\} is affiliated with is \{tail\}. & 170,931 \\
    producer & The producer of \{head\} is \{tail\}. & 169,628 \\
    formerTeam & \{head\} used to play for \{tail\} team. & 151,374 \\
    family & \{head\} belongs to the \{tail\} family. & 148,818 \\
  currentMember & The current member of \{head\} is \{tail\}. & 148,739 \\
    battle & \{head\} participated in the following battles: \{tail\}. & 148,188 \\
    nationality &  The nationality of {head} is {tail}. & 147,525 \\ 
    director & The director of \{head\} is \{tail\}. & 145,621 \\
    associatedBand & The band associated with \{head\} is \{tail\}. & 135,597 \\
    associatedMusical& The musical artist associated with \{head\}   & 135,582 
    \\$\;\hookrightarrow$ Artist &$\;\hookrightarrow$  in the music industry is \{tail\}. & \\
    class & The class of \{head\} is \{tail\}. & 127,837 \\
    order & The order of \{head\} is \{tail\}. & 123,626 \\
    \bottomrule
    \end{tabular}}
    \caption{Relations templates.}
\label{tab:relation_templates}
\end{table}

\paragraph{Relation templates}
We use the relation templates to create queries for evaluating the models.
These templates are first generated by GPT with Web API in a few-shot manner, then manually curated to ensure the quality of the templates. We refer to Figure \ref{fig:relation_template_prompt} for the prompt used
for generating the relation templates. The prompt is designed to ask the model to generate a query for a given relation type. The model is asked to generate a query that can be used to judge the factuality of the relation type. We then manually curate the generated templates to ensure the quality of the templates. 
We refer to Table \ref{tab:relation_templates} for the curated relation templates.  Due to the large number of relation types in the DBpedia knowledge graph, we only showcase a few relation templates in the table, these templates are the most common relation types in the knowledge graph, sorted by the number of triples associated with the relation type.
 We can see that the relation templates are comprehensive and cover a wide range of topics. This can be seen as a source of multiple-domain knowledge for evaluating the LLMs.

\paragraph{Data and Model}
 We download the DBpedia data dump from \href{https://www.dbpedia.org/}{https://www.dbpedia.org/}. We use the turtle format of the DBpedia knowledge graph. 
 We directly use the LLaMA 2 and Gemma from the Hugging Face model hub. The model cards are
 \href{https://huggingface.co/meta-llama/Llama-2-7b-chat-hf}{meta-llama/Llama-2-7b-chat-hf}, \href{https://huggingface.com/meta-llama/Llama-2-13b-chat-hf}{ meta-llama/Llama-2-13b-chat-hf}, \href{https://huggingface.co/meta-llama/Llama-2-70b-chat-hf}{meta-llama/Llama-2-70b-chat-hf}, \href{https://huggingface.co/gemma-team/gemma-2b-chat-hf}{gemma-team/gemma-2b-chat-hf}, and \href{https://huggingface.co/gemma-team/gemma-7b-chat-hf}{gemma-team/gemma-7b-chat-hf}.

\begin{table}[t]
    \centering
    \begin{tabular}{l|p{4.7in}}
    \toprule
    \textbf{Model} & \textbf{Instruction} \\ \midrule
    LLaMA 2 &   Below is an instruction that describes a task, paired with an input that provides further context. Write a response that appropriately completes the request.\textbackslash n\textbackslash n \#\#\# Instruction:\textbackslash n You are given a statement. You are asked to judge whether the statement is true or false. Answer 'Yes, the statement is true.' if you know the statement is true. Answer 'No, the statement is false.' if you know the statement is false. Otherwise, answer 'I don't know.'\textbackslash n\textbackslash n\#\#\# Input: {\bf Input} \textbackslash n\textbackslash n\#\#\# Response:\textbackslash n\textbackslash n 
                \\ \hline
    
    Gemma & start\_of\_turn\textgreater user  Below is an instruction that describes a task, paired with an input that provides further context. Write a response that appropriately completes the request.\textbackslash n\textbackslash n\#\#\# Instruction:\textbackslash n You are given a statement. You are asked to judge whether the statement is true or false. Answer 'Yes, the statement is true.' if you know the statement is true. Answer 'No, the statement is false.' if you know the statement is false. Otherwise, answer 'I don't know.'\textbackslash n\textbackslash n\#\#\# Input: {\bf Input} \textless end\_of\_turn\textgreater\textless start\_of\_turn\textgreater model\textbackslash n\textbackslash n The answer is " \\
    
    \bottomrule
    \end{tabular}
    \caption{Instruction used for creating queries.} %
    \label{tab:instruction} %
\end{table}

\paragraph{Instruction used for the LLaMA and Gemma models}
We report the instructions used for creating queries for the LLaMA and Gemma models. The instruction is designed to ask the model to judge whether the statement is true or false. We refer to Table \ref{tab:instruction} for the instruction used for creating queries. We use the same instruction for both the LLaMA and Gemma models with little modification to adjust the model's instruction format. With this instruction, the most frequent responses of LLMs are \textit{Yes, the statement is true}, \textit{No, the statement is false}, and \textit{I don't know}, with some variations on the suffix, mainly explaining the reason for the answer. This is what we expect from the LLMs when using a judge model (or the first-token logit as well), since the judge model doesn't use the LLM's response, but the hidden state of the LLM, which makes the consistency of the response format important.

\end{document}